%% file: main.tex
\documentclass[10pt,twocolumn,letterpaper]{article}

\usepackage{cvpr}      
\usepackage{graphicx}  
\usepackage{natbib}    
\usepackage{booktabs}  
\usepackage{tcolorbox}
\tcbuselibrary{breakable}
\tcbuselibrary{skins}

\usepackage{setspace}     
\usepackage{amssymb}      
\usepackage{amsmath}      
\usepackage{tabularx}
\usepackage{algorithm}
\usepackage{amsmath}
\usepackage{amssymb}
\usepackage{booktabs}
\usepackage{float}
\usepackage{amsfonts}
\usepackage{xcolor}
\usepackage{amsmath}
\usepackage{CJKutf8}
\usepackage{algorithmic}
\usepackage{tabularray}
\usepackage{tcolorbox}
\usepackage{listings}
\usepackage{xcolor} 
\usepackage{amsmath}
\usepackage{makecell}
\usepackage{tabularx}
\usepackage{booktabs}    
\usepackage{array}       
\usepackage{multirow}    
\usepackage{makecell}    
\usepackage{graphicx}    
\usepackage[table]{xcolor} 
\usepackage{colortbl}    
\usepackage[section]{placeins}
\usepackage{booktabs}
\usepackage{float}
\usepackage[table]{xcolor}
\definecolor{orange}{RGB}{255,127,0} 
\input{preamble}
\definecolor{cvprblue}{rgb}{0.21,0.49,0.74}
\usepackage[pagebackref,breaklinks,colorlinks,allcolors=cvprblue]{hyperref}

\def\paperID{1653} 
\def\confName{CVPR}
\def\confYear{2026}

\title{Towards Efficient Multimodal Unified Reasoning Model via Model Merging}

\author{Qixiang Yin\textsuperscript{1,5}\footnotemark[1]~~, 
Huanjin Yao\textsuperscript{2}\footnotemark[1]~~, 
Jianghao Chen\textsuperscript{5}, 
Jiaxing Huang\textsuperscript{2}, Zhicheng Zhao\textsuperscript{1,3,4}\footnotemark[2]~~, 
Fei Su\textsuperscript{1,3,4}\footnotemark[2] \\
\\
\textsuperscript{1} Beijing University of Posts and Telecommunications\quad
\textsuperscript{2} Nanyang Technological University \\
\textsuperscript{3} Beijing Key Laboratory of Network System and Network Culture\\
\textsuperscript{4} Key Laboratory of Interactive Technology and Experience System, Ministry of Culture and Tourism \\
\textsuperscript{5}~Zhongguancun Academy, Beijing, China\\
\texttt{\{buptyqx, zhaozc, sufei\}@bupt.edu.cn} 
        }

\begin{document}
\maketitle

\renewcommand{\thefootnote}{\fnsymbol{footnote}}
\footnotetext[1]{Equal contribution.}
\footnotetext[2]{Corresponding authors.}
\renewcommand*{\thefootnote}{\arabic{footnote}}

\begin{abstract}
Although Multimodal Large Language Models (MLLMs) have demonstrated remarkable capabilities across diverse tasks, they encounter challenges in terms of reasoning efficiency, large model size and overthinking. However, existing lightweight MLLMs lack the capability to balance high efficiency and performance at a small scale. To this end, we propose Tiny-R1V, a novel lightweight 3B model that achieves faster inference and higher accuracy via a two-stage optimization, while unifying multimodal reasoning across multiple tasks with fewer inference tokens. In the first stage, Tiny-R1V introduces Length-Informed Relative Policy Optimization (LIPO), a new reinforcement learning method, to train each reasoning model, including mathematical reasoning, chart reasoning, and OCR capability. The LIPO dynamically adjusts the advantages of responses within groups by prioritizing concise yet high-quality responses to encourage the generation of shorter and more accurate responses. In the second stage, we propose Adaptive Model Merging (AMM), a training-free model merging method that merges multiple specialist models into a unified architecture. Specifically, AMM adaptively adjusts the weights of task vectors via a novel gradient projection regularization loss function, thus mitigating redundant conflicts between them. Extensive evaluations on ten widely-used reasoning benchmarks covering mathematics, structured data (charts, tables, documents), OCR, and general capabilities showcase the superior performance of Tiny-R1V, enabling lightweight models to excel in diverse multimodal reasoning tasks. Code will be available at \href{https://github.com/buptyqx/Tiny-R1V}{https://github.com/buptyqx/Tiny-R1V}
\end{abstract}

\input{src/1_introduction}

\input{src/2_related_work}

\input{src/3_method}

\input{src/4_experiment}

\section{Acknowledgments}
This work is supported by the BUPT Innovation and Entrepreneurship Support Program (2025-YC-T033).
\input{src/appendix}
{
    \small
    \bibliographystyle{ieeenat_fullname}
    \bibliography{main}
}


\end{document}

%% file: src/1_introduction.tex
\section{Introduction}
\begin{figure}[htbp]
\centering
\includegraphics[width=\columnwidth]{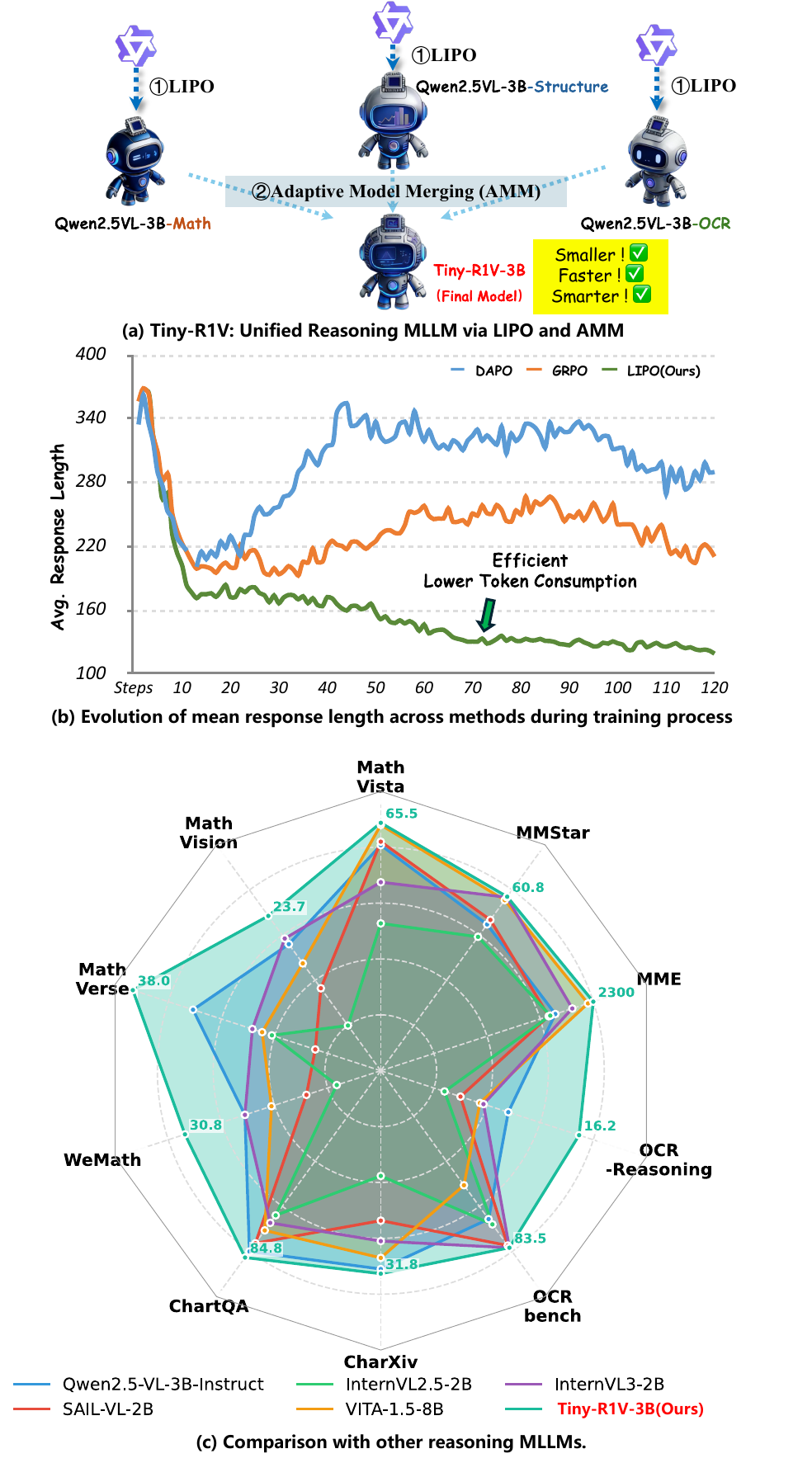} 

   \caption{(\textbf{a}) Two-stage framework for training lightweight MLLMs for unified reasoning tasks. (\textbf{b}) The average response length of GRPO~\citep{shao2024deepseekmath}, DAPO~\citep{yu2025dapo} and LIPO(Ours) on the training set during the RL training process. (\textbf{c}) Tiny-R1V achieves the state-of-the-art performance on a broad range of multimodal reasoning tasks compared with other open source models.}
   \label{fig:firstpage}
\end{figure}

Multimodal Large Language Models (MLLM)~\citep{bai2025qwen2_5_vl,hurst2024gpt4o,team2023gemini,yao2024minicpmv} have shown powerful capabilities in extensive applications across different tasks. However, MLLMs still encounter several challenges in terms of reasoning ability. On the one hand, the improvement of the models' reasoning ability is constrained by scaling laws~\citep{kaplan2020scaling_law}, which limit the extent to which model reasoning performance can be significantly enhanced when the parameter scale is small. In particular, as model size increases, there is often a diminishing return on reasoning efficiency, thus creating a trade-off between model size and inference speed. On the other hand, these models also face challenges such as the high cost of large-scale training, which further limits the ability to explore larger models. Additionally, there are issues in achieving a balance when jointly training different reasoning tasks—where optimizing performance across multiple tasks may hinder specialization in any single task, thus complicating task integration and model efficiency.

The recent success of Reinforcement Learning (RL) in Large Language Models (LLMs), such as Kimi-K1.5~\citep{team2025kimik1_5} and DeepSeek-R1~\citep{guo2025deepseekr1} have demonstrated its potential in motivating long chain-of-thought (CoT) reasoning ability~\citep{yeo2025demystifying, chen-etal-2025-lr2bench} through rule-based Group Relative Policy Optimization (GRPO)~\citep{shao2024deepseekmath}, enabling LLMs to handle complex reasoning tasks. However, due to long CoT, the existing reasoning models will inevitably incur high inference costs and suffer from the issue of over-thinking~\citep{sui2025stopoverthinking,chen2024not}, which restricts their deployment in real-time or resource-constrained scenarios. Therefore, to reduce the redundancy of long-CoT, some efficiency-oriented methods are proposed through pruning~\citep{luo2025o1_pruner} or compression~\citep{chen2024not}, while they fundamentally overlook the potential for greater gains from shorter reasoning CoT.


Moreover, Moreover, in the field of traditional vision tasks and LLMs~\citep{akiba2025evolutionary,ilharco2022task_arithmetic,cheng2025wudi}, model merging~\citep{akiba2025evolutionary,ilharco2022task_arithmetic,cheng2025wudi} techniques have emerged as a powerful training-free approach to combine multiple specialist models into a unified architecture, which retains the capabilities of each expert model while improving the unified understanding ability. However, in terms of reasoning capability, existing model merging methods~\citep{ilharco2022task_arithmetic,yadav2023ties_merging,yu2024dare} fail to adequately preserve the strengths of each expert model, often leading to substantial performance degradation during the merging process and resulting in an inability to properly balance different tasks. Hence, corresponding to the above progress, a critical question emerges: \textbf{\textit{Is it possible to devise a novel paradigm that empowers lightweight models to achieve efficient and accurate reasoning across diverse tasks?}}


In this paper, as shown in Figure~\ref{fig:firstpage}(a), we propose a lightweight 3B model named Tiny-R1V, which achieves both faster inference speed and more accurate reasoning performance through a joint two-stage learning framework. Overall, stage 1 is dedicated to training the model to use efficient CoT reasoning, while stage 2 focuses on balancing the tasks to optimize performance across multiple reasoning capabilities. Specifically, in the first stage, for RL, we provide large-scale training data encompassing various tasks such as geometry, charts, tables, and OCR. For each task, we collect public datasets containing at least 10k samples to ensure effective post-training via RL. Building upon this, we design a novel Length-Informed Relative Policy Optimization (LIPO) to dynamically adjust the advantages of responses within the group, which aims to reduce the advantages of longer responses and increase those of shorter ones among responses that have approximately equivalent rewards. In this manner, as shown in Figure~\ref{fig:firstpage}(b), each response is constrained within a valid range, while ensuring that the model can output more accurate responses in the form of concise answers as much as possible.

In the second stage, we propose a novel model merging method, namely Adaptive Model Merging (AMM), aiming to enhance the optimization of task vectors (i.e., parameter changes between post-training models and the base model). Specifically, AMM adaptively adjusts the weights of task vectors through inherent task importance parameters $\alpha$ and dynamic state parameters $\beta$, and robustly optimizes the merged vectors via the gradient projection regularization loss function. Meanwhile, AMM enables the integration of multiple MLLMs without requiring additional data for training. Furthermore, as shown in Figure~\ref{fig:firstpage}(c), the proposed merging method effectively consolidates inputs from diverse tasks and outperforms state-of-the-art (SOTA) models trained on mixed training data. The main contributions of this work are summarized as follows: 
\begin{enumerate}
    \item We propose a two-stage framework for training lightweight multimodal models for reasoning tasks, and then construct Tiny-R1V with only 3B parameters.
    \item We introduce Length-Informed Relative Policy Optimization (LIPO), which dynamically adjusts the inter-group response advantages, minimizing the number of response tokens while ensuring the accuracy of the answer.
    \item We design a novel Adaptive Model Merging (AMM) , which not only retains the unique advantages of each model, but also reduces the redundant interference between models during the model merge process.
    \item Extensive experiments on ten MLLM reasoning benchmarks derived from four different tasks demonstrate the superiority of our proposed Tiny-R1V.
\end{enumerate}

%% file: src/2_related_work.tex
\section{Related Work}

\subsection{Reinforcement Learning for MLLMs}

RL~\citep{cao2024surveyRL} has been proven to be a key technology for enhancing the reasoning capabilities of LLMs. Early research mainly focused on Reinforcement Learning from Human Feedback (RLHF)~\citep{bai2022RLHF,yu2024rlhfv}. However, RLHF's reliance on high-quality manual annotations limits its scalability in larger-scale scenarios. To address this, various rule-based reward functions~\citep{guo2025deepseekr1, chen2025ace} have been proposed to provide reliable rewards for reinforcement learning without complex human annotations. Inspired by this, recent studies have begun to introduce reinforcement learning into the multimodal~\citep{yao2025mmreason,yao2024mulberry,zhang2025r1vl,yao2025r1sharevl} domain to boost model performance in complex visual reasoning tasks. Existing works~\citep{yang2025r1onevision,liu2025segzero} have designed fine-grained rule-based reward functions from various perspectives, such as answer correctness~\citep{peng2025skyworkr1v}, reasoning chain completeness~\citep{zhang2025r1vl}, and visual consistency~\citep{huang2025visionr1}, which effectively improve the accuracy and robustness of MLLMs' reasoning capability.

\subsection{Model Merging}
Model merging~\citep{yang2024model_merge_survey} emerges as a cost-effective and flexible strategy to enable the integration of capabilities from multiple expert models without additional training. The training-free weight merging method assumes that all expert models share the same initialization parameters, and directly applies strategies such as linear interpolation~\citep{wortsman2022model_average,ilharco2022task_arithmetic}, sparsification~\citep{yadav2023ties_merging,yu2024dare}, and low-rank optimization to the parameters in the weight space~\citep{choi2024revisiting,cheng2025wudi,wei2025wudiv2,miyano2025adaptive_lora_merge,zhang2025lori} to achieve parameter merging. Meanwhile, inspired by the idea of the Mixture-of-Experts (MoE) architecture, the dynamic routing merging~\citep{tang2024merging_moe,muqeeth2023soft_merging} method dynamically calculates the weights of each expert model based on input samples during the inference phase and generates combination coefficients in real-time through a lightweight routing mechanism.

%% file: src/3_method.tex
\section{Method}

This section first provides the preliminaries, then introduces the key techniques of the proposed Tiny-R1V framework, namely \textbf{L}ength-\textbf{I}nformed Relative \textbf{P}olicy \textbf{O}ptimization (\textbf{LIPO}) and \textbf{A}daptive \textbf{M}odel \textbf{M}erging (\textbf{AMM}).

\subsection{Preliminaries}

The GRPO~\citep{shao2024deepseekmath}framework initially leverages a MLLM to initialize both a policy model $\pi_{\theta}$ and a reference model $\pi_{\text{old}}$. For a given image-text pair $(\mathcal{I},\mathcal{T})$, the reference policy model $\pi_{\theta_{\mathrm{old}}}$ generates $G$ responses $\{{o}_1, {o}_2, ..., {o}_G\}$. A group-based reward function then computes the corresponding rewards $\{R_1, R_2, ..., R_G\}$, which are subsequently utilized to estimate the advantage $\hat{A}{i}$ for each response within the group, $\hat{A}_{i} = (R_i -\text{mean}\left(\{R_i\}_{i=1}^{G}\right))/\text{std}\left(\{R_i\}_{i=1}^{G}\right)$. Then, GRPO employs a clipped objective with a KL penalty term: 
\begin{equation}
\begin{aligned}
&\mathcal{J}_{\mathrm{GRPO}}(\theta) = \mathbb{E}_{(\mathcal{I},\mathcal{T})\sim p_{\mathcal{D}}, o\sim\pi_{\theta_\text{old}}(\cdot|\mathcal{I},\mathcal{T})} \\
&\quad \Biggl[ \frac{1}{n}\sum_{i=1}^{n} \min\ \!\Biggl(\frac{\pi_{\theta}(o_i \mid \mathcal{I},\mathcal{T})}{\pi_{\theta_{\mathrm{old}}}(o_i \mid \mathcal{I},\mathcal{T})}\hat{A}_i, \mathrm{clip}\ \!\Bigl(\frac{\pi_{\theta}(o_i \mid \mathcal{I},\mathcal{T})}{\pi_{\theta_{\mathrm{old}}}(o_i \mid \mathcal{I},\mathcal{T})}, \\
&\quad\quad 1-\epsilon,\,1+\epsilon\Bigr)\hat{A}_i - \beta D_{\mathrm{KL}}\left(\pi_{\theta}\mid\mid \pi_{\mathrm{ref}}\right)\Biggr) \Biggr]
\end{aligned}
\end{equation}

\textbf{WUDI Model Merging.} Given a pre-trained base model $\theta_0$ and $P$ task-specific models $\{\theta_i\}_{i=1}^P$, the task vector for each task $i$ is defined as the parameter difference between the task-specific model and the base model, $\boldsymbol{\tau}_i = \theta_i - \theta_0$.
WUDI-merging~\citep{cheng2025wudi} aims to minimize the interference between the merged task vector and each individual task vector, as $\boldsymbol{\tau}_{m, l}-\boldsymbol{\tau}_{i, l} $  for task $ i $ at layer $l$. Intuitively, it encourages the merged task vector to retain the key information of each task vector in its own direction, thereby reducing redundancy and conflict between tasks. For each layer $l$, the merged task vector is iteratively refined over $\mathcal{N}$ steps using a specialized loss function:
\begin{align}
\min _{\boldsymbol{\tau}_{m, l}} \mathcal{L}_{l}=\sum_{i=1}^{n} \frac{1}{\left\|\boldsymbol{\tau}_{i, l}\right\|_{F}^{2}}\left\|\left(\boldsymbol{\tau}_{m, l}-\boldsymbol{\tau}_{i, l}\right)\left(\boldsymbol{\tau}_{i, l}\right)^{\top}\right\|_{F}^{2} 
\label{eq:wudi_merge}
\end{align}
where $\|\cdot\|_F$ denotes the Frobenius~\citep{bottcher2008frobenius} norm of a matrix.

\begin{figure*}[htbp]
\centering
\includegraphics[width=0.98\textwidth]{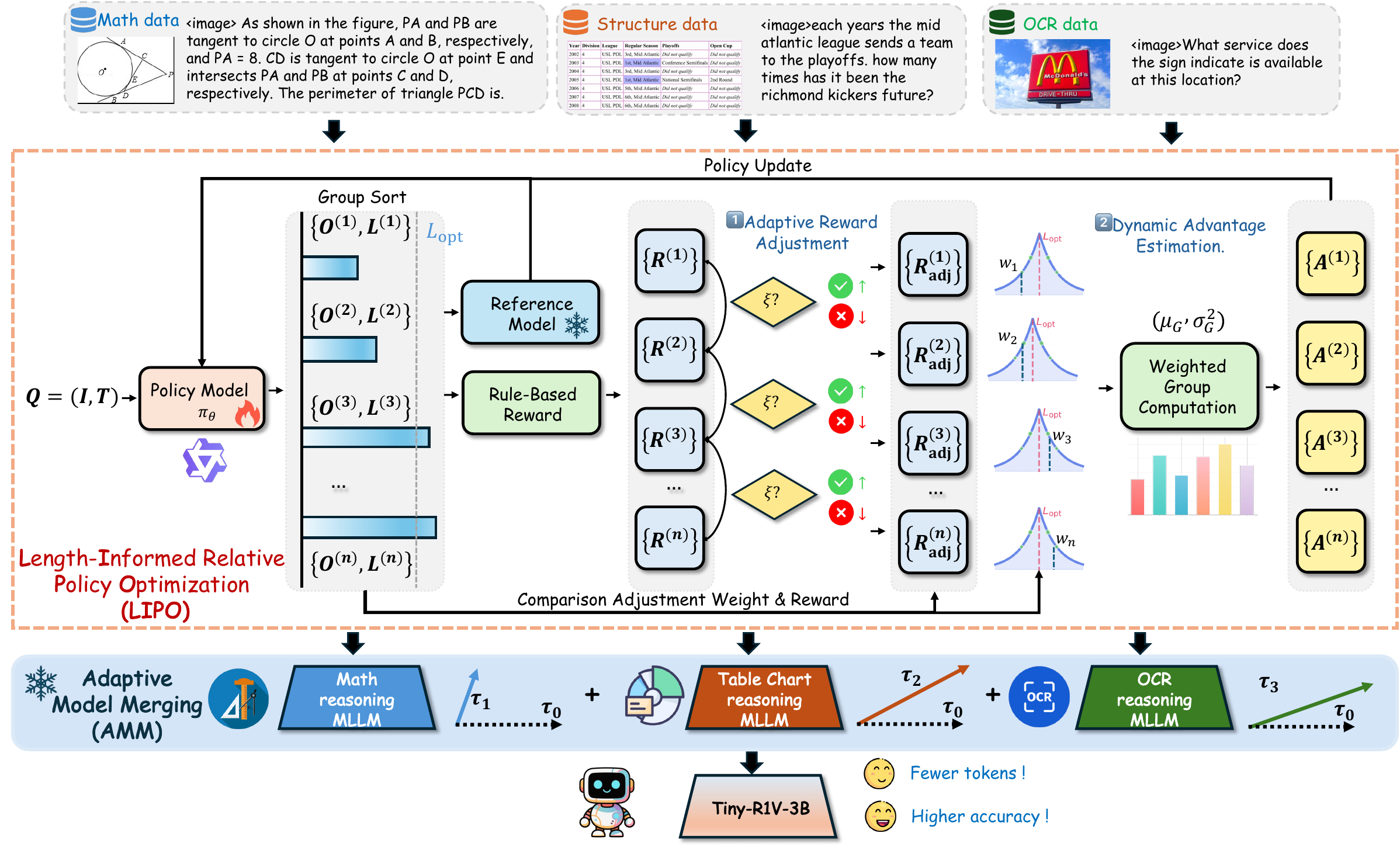} 
\caption{\textbf{Overview of the proposed Tiny-R1V}. Tiny-R1V employs Length-Informed Relative Policy Optimization (LIPO) and Adaptive Model Merging (AMM). In the first stage, Tiny-R1V trains three expert models separately using LIPO, which dynamically adjusts the advantages between groups to minimize the number of response tokens while ensuring the accuracy of the answer. In the second stage, Tiny-R1V merges the three models using AMM, determines the dynamic weights of model parameters, and reduces parameter conflicts, resulting in the final Tiny-R1V-3B model.
}
\label{fig:LIPO}
\end{figure*}

\subsection{\textbf{LIPO}: \textbf{L}ength-\textbf{I}nformed Relative \textbf{P}olicy \textbf{O}ptimization} 
To achieve lightweight model reasoning, relieve the problem of overthinking, and favor simple yet correct responses, we propose \textbf{L}ength-\textbf{I}nformed Relative \textbf{P}olicy \textbf{O}ptimization (\textbf{LIPO}), a novel online MLLM reinforcement learning framework. It rewards shorter and correct responses through adaptive reward adjustment and dynamic advantage estimation. 

For a given question $q \in \mathcal{Q} $ and $k$ responses generated by the model $\mathcal{O} = \{o^{(1)}, o^{(2)}, \dots, o^{(k)}\}$, the length of each response is $L = \{L^{(1)}, L^{(2)}, \dots, L^{(k)}\}$, a response $o^* \in \mathcal{O}$ is Pareto optimal~\citep{ngatchou2005pareto} if and only if there does not exists any other response $o \in \mathcal{O}$ such that at least one of the following conditions holds  $Q(o) \geq Q(o^*)\ \text{or} \ L(o) \leq L(o^*)$, within the feasible solution set in the quality-length ($Q$-$L$)  space $\mathcal{F} = \{(Q(o), L(o)) \mid o \in \mathcal{O}\}$.

\textbf{Adaptive Reward Adjustment.} The core motivation is to correct the $Q$-$L$ imbalance problem of the traditional reward mechanism,  explicitly favoring concise yet high-quality responses while ensuring meaningful reasoning length. Responses within a group are sorted by length in ascending order as $  L^{(1)} \leq L^{(2)} \leq \cdots \leq L^{(k)}  $. For a pair of adjacent responses to qualify for reward adjustment, $ (L^{(i)}, R^{(i)}) $ and $ (L^{(i+1)}, R^{(i+1)}) $, they must satisfy a trigger condition $\xi(o^{(i)}, o^{(i+1)})$, defined as $\xi(o^{(i)}, o^{(i+1)}) = \left( |R^{(j)} - R^{(i)}| < \eta \right)
\land \left( L^{(i+1)} - L^{(i)} > 0 \right) \land \left( L^{(i)} \geq L_{\min} \right)$. This condition encodes three critical constraints:$ |R^{(j)} - R^{(i)}| < \eta$ ensures that the difference in reward values between two adjacent responses is within a certain threshold $ \eta $, indicating that their qualities are relatively close. $ L^{(i+1)} - L^{(i)} > 0 $ and $ L^{(i)} \geq L_{\min} $ ensure that responses with appropriate length are selected through comparison. Based on this trigger condition, we adjust the reward for the shorter response $R^{(i)}$ as follows:
\begin{align}
R_{\text{adj}}^{(i)} & = \begin{cases}  R^{(i)} (1 + \alpha) & \xi \land L^{(i)} < L_T \\ 
R^{(i)} \left(1 + \alpha\omega )\right) & \xi \land L^{(i)} \geq L_T \\ 
R^{(i)} & \text{otherwise} \end{cases}
\end{align}
where, $\alpha \in (0, 0.2]$ is a base enhancement factor, and $\omega $ is a decay term that modulates the enhancement, defined as $\omega (o^{(i)}, o^{(i+1)}) = \max\left(0, 1 - \frac{L^{(i)} - L_T}{L^{(i+1)} - L_T + \epsilon}\right)$. For responses significantly shorter than a threshold $L_T$, we apply the full enhancement of $\alpha$ to strongly incentivize such concise outputs. For responses whose lengths are longer than $L_T$, when $L^{(i)}$ approaches $L^{(i+1)}$, it means that the shorter response is only slightly shorter than the longer one. At this time, $\omega$ tends to 0, thus reducing the enhancement. In summary, this design achieves a smooth transition and balances the reward signal and the simplicity of the response.

\textbf{Dynamic Advantage Estimation.} We propose a dynamic advantage estimation to normalize the reward by calculating the advantage of each response relative to the ideal length within the group. The ideal length of each response group is defined as $L_{\text{opt}}  = \max(2L_{\min},\text{median}(L^{(i)}))$. By taking the maximum of these two values, $L_{\text{opt}}$ ensures that the optimal length is both practical and representative of the group's characteristics. Then, the weight $  w_i$ for each response is computed as:
\begin{align}
w_i = \exp\left(-\phi  |L^{(i)} - L_{\text{opt}}|\right)
\end{align}
where $\phi$ is the scaling factor and the weight $w_i$ reflects the proximity of each response's length to the optimal length $  L_{\text{opt}}  $. Responses that are closer to $  L_{\text{opt}}  $ will be assigned higher weights, emphasizing their importance in the advantage estimation process. The dynamic advantage $  A^{(i)}  $ is then calculated using the following set of equations:
\begin{align}
\mu_G = \dfrac{\sum_{i \in G} w_i R_{\text{adj}}^{(i)}}{\sum_{i \in G} w_i} \quad& \sigma_G^2 = \dfrac{\sum_{i \in G} w_i (R_{\text{adj}}^{(i)} - \mu_G)^2}{\sum_{i \in G} w_i} \\
    A^{(i)} &= \dfrac{R_{\text{adj}}^{(i)} - \mu_G}{\sigma_G + \epsilon} 
\end{align}
where, $  \mu_G $ is the weighted average of the adjusted rewards, $  \sigma_G^2 $ is the weighted variance. This dynamic advantage estimation helps the model better understand the relative quality of each response within the group, guiding it to generate more optimal answers. 

\textbf{Reward function.} Consistent with GRPO, LIPO also adopts rule-based rewards, with exact matching rewards for maths tasks and structured data (table, chart, document) tasks. For OCR tasks, a more continuous Levenshtein Distance~\citep{yujian2007ld} is utilized, where the reward is calculated by comparing the edit distance between the response string and the target string.

\begin{table*}[htbp]
  \renewcommand{\arraystretch}{1}
  \caption{Capability merging results on Qwen2.5-VL-Instruct (RL post training) across multiple tasks. 
  The best and second best average results are highlighted in boldface and underlined respectively. 
  $^\ast $ denotes evaluation on official open-source weights using VLMEvalKit~\citep{duan2024vlmevalkit}.}
  \label{tab:qwen2_5vl_main}
  \centering
  \setlength{\tabcolsep}{2pt}
  \resizebox{\textwidth}{!}{
\begin{tabular}{l|cccc|cc|cc|cc|c}
  \toprule
  \multirow{2}{*}{\textbf{Methods}}      & \multicolumn{4}{c|}{\textbf{Math}} & \multicolumn{2}{c|}{\textbf{Table \& Chart \&Doc}} & \multicolumn{2}{c|}{\textbf{OCR}} & \multicolumn{2}{c|}{\textbf{General}} & \multirow{3}{*}{\textbf{Avg.}}                                              \\ \cmidrule{2-11}
 &  \thead{MathVista \\(mini)} & \thead{MathVision \\(full)} & \thead{MathVerse \\(O Vision)} & \thead{WeMath \\(Strict)} & \thead{ChartQA\\(test Avg.) }& \thead{CharXiv\\(RQ)} & \thead{OCRbench} & \thead{OCR-\\Reasoning } & \thead{MME \\ (sum) } & \thead{MMStar} &      \\ \midrule
Qwen2.5-VL-3B-Instruct~\citep{bai2025qwen2_5_vl} &  62.3 &  21.2  &  31.2 & 	22.9 & 84.0 & 31.3 &  79.7 & 12.2 & 2157 & 55.9 & 47.8 \\ \midrule
\multicolumn{12}{c}{\textbf{\textit{Individual models}}} \\ \midrule

\textbf{Individual Math Model-LIPO (Ours)}& 66.5 & 23.3 & 35.8 & 30.7 & 83.9 & 32.0 & 73.5 & 16.6& 2304 & 58.5 & 50.3  \\
\textbf{Individual  Structure Model-LIPO (Ours)}& 61.9 & 22.0 & 33.1 & 23.6 & 85.8 & 34.5& 77.6 & 14.5 & 2302 & 58.7 & 49.4 \\
\textbf{Individual OCR Model-LIPO (Ours)} & 57.9 & 20.0 & 31.9 & 26.3 & 83.5 & 30.4 & 84.6 & 13.3 & 2254 & 57.4 & 48.6 \\ \midrule
\multicolumn{12}{c}{\textbf{\textit{Comparison with other merging methods}}} \\ \midrule
Task Arithmetic~\citep{ilharco2022task_arithmetic} & 62.3 & 21.6 & 28.6 & 26.6 & 78.2 & 30.1 & 77.3 & 12.0 & 2044 & 56.6 &  46.7\\
TA+Dare~\citep{yu2024dare} & 62.2 & 21.8 & 28.5 & 26.6 & 79.6 & 31.1 & 76.2 & 12.3 & 2034 & 56.8 &  46.8 \\
TIES Merging~\citep{yadav2023ties_merging} & 61.6 & 22.2 & 33.2 & 28.4 & 83.9 & 31.3 & 78.0 & 13.2 & 2298 & 58.0 & 49.2 \\
TIES+Dare~\citep{yu2024dare} & 62.6 & 23.0 & 33.6 & 28.9 & 84.0 & 31.6 & 78.4 & 13.2 & 2312 & 58.2 &  49.6 \\
TSV Merging~\citep{gargiulo2025tsv} & 61.5 & 23.5 & 34.9 & 30.6 & 85.0 & 31.4 & 79.6 & 14.0 & 2269 & 59.5 & 50.1 \\
Iso-C Merging~\citep{marczak2025iso_c} & 62.1 & 22.2 & 34.9 & 29.5 & 84.9 & 32.2 & 80.0 & 13.3 & 2286 &  57.6 & 49.8  \\
WUDI Merging~\citep{cheng2025wudi} & 63.5 &  23.3 & 37.8 & 30.3 & 85.0 & 32.4 &  81.5 & 15.2 & 2262 & 58.7 & 50.8 \\
WUDI Merging v2 ~\citep{wei2025wudiv2}& 61.8 & 23.4 & 34.7 & 31.4 & 85.0 & 32.2 & 79.8 & 15.0 & 2269 & 59.5 & 50.4 \\ \midrule
\multicolumn{12}{c}{\textbf{\textit{Comparison with mixture training}}} \\ \midrule
Mixture Training & 64.5 & 23.2 & 34.2 & 30.0 & 84.8 & 32.0 & 82.0 & 12.8 & 2269 & 59.5 & 50.4 \\  \midrule
\multicolumn{12}{c}{\textbf{\textit{Comparison with other models}}} \\ \midrule
SAIL-VL-2B~\citep{dong2025sail} & 62.8 & 	17.3 & 17.4 &  14.8 &  82.9 & ~~26.1$^*$ & 83.2 & ~~9.5$^*$ & ~~2132$^*$& 56.7 & 44.7\\
InternVL2.5-2B~\citep{chen2024internvl2_5} &  51.1 & 14.0 & 22.3 & 10.8 & 79.2 & 21.3 & 80.4 & ~~8.6$^*$ & 	2138 & 53.7 &  41.8\\ 
InternVL3-2B ~\citep{zhu2025internvl3} & 57.0 & 21.7 & 24.5 & 	22.9 & 80.2 & 28.3 & 83.5 & 10.8 & 2221 & 60.7 &46.9  \\
VITA-1.5-8B~\citep{fu2025vita1_5} & ~~65.2$^*$ & 19.5 & 23.4 & 	19.4 & ~~81.2$^*$& ~~30.1$^*$ &75.2 & ~~10.6$^*$& ~~2280$^*$ & 60.2 & 46.6\\

\hline

\rowcolor{gray!15} \textbf{Tiny-R1V-3B (Ours)} & \textbf{65.5}& \textbf{23.7} & \textbf{38.0} & \textbf{30.8} & \textbf{85.3} & \textbf{32.4} & \textbf{82.5} &\textbf{16.2} &\textbf{ 2291 }& \textbf{59.5} & \textbf{51.6}\\

\bottomrule
\end{tabular}
  }
\end{table*}

\subsection{\textbf{AMM}: \textbf{A}daptive \textbf{M}odel \textbf{M}erging}

After obtaining multiple expert models, an important challenge lies in how to fully integrate the capabilities of each model. WUDI Merging~\citep{cheng2025wudi} considers only the task vectors of the models, even though different expert models contribute with unequal weights along the task-vector direction. Moreover, during the merging process, the gradient directions contributed by different expert models across layers are also misaligned, which makes it difficult to effectively balance multi-task performance.

We propose a dual-weight adaptive mechanism to achieve refined control of task vectors by introducing two weight parameters, $\alpha$ and $\beta$. The $\alpha_{i,l} = \frac{\|\boldsymbol{\tau}_{i,l}\|_F^2}{\sum_{j=1}^{n} \|\boldsymbol{\tau}_{j,l}\|_F^2}$ parameter measures the inherent importance of tasks based on the norm of the task vector to ensure that key tasks are not neglected. The $\beta_{i, l}^{n}$ parameter is dynamically adjusted based on the current merge state. It is used to measure the compatibility of tasks with the merge vector and focus on the tasks that match the current merge result best.
\begin{align}
    \beta_{i,l}^n = \exp\left(-\gamma \cdot \frac{\|(\boldsymbol{\tau}_{m,l}^{n-1} - \boldsymbol{\tau}_{i,l})\boldsymbol{\tau}_{i,l}^\top\|_F^2}{\|\boldsymbol{\tau}_{i,l}\|_F^2}\right)
\label{con:beta}
\end{align}
where $\gamma$ is the scaling factor. The calculation of $\beta_{i, l}^{n}$ is based on the merged vector $\boldsymbol{\tau}_{m, l}^{n-1}$ after the $(n-1)$th iteration, so that it will be dynamically adjusted according to the latest state of the merged vector. As the iteration proceeds, the merged vector $\boldsymbol{\tau}_{m, l}$ gradually converges, and $\beta_{i, l}^{n}$ will also tend to be stable, thus the merged vector can retain the key information of multiple tasks at the same time.

\begin{algorithm}[htbp]

\caption{AMM: Adaptive Model Merging}

\label{algorithm1}

\begin{algorithmic}[1]
\setstretch{0.8}
\STATE\textbf{Input:} Parameters $\boldsymbol{\theta}$; task vectors $\mathcal{T}=\{\boldsymbol{\tau}_{i,l}\}_{i=1}^\mathcal{P}$; steps $\mathcal{N}$; learning rate $\zeta$; regularization strength $\gamma$; gradient regularization strength $\lambda$

\STATE\textbf{Output:} Merged parameters $\boldsymbol{\theta_m}$.

\STATE $\triangleright$ \textit{Initialize with weighted sum ($\star$)}

\FOR{linear layer $l \in\{1,\cdots,\Psi\}$}

\STATE Compute $\alpha_{i,l},\beta_{i, l}^{0}$

\STATE $\boldsymbol{w}_{i,l}^0 = \alpha_{i,l} \cdot \beta_{i,l}^0 / \sum_{j=1}^\mathcal{P} (\alpha_{j,l} \cdot \beta_{j,l}^0)$
\STATE $\boldsymbol{\tau}_{m,l}^0 = \sum_{i=1}^\mathcal{P} \boldsymbol{w}_{i,l}^0 \cdot \boldsymbol{\tau}_{i,l}$ 

\ENDFOR

\FOR{linear layer $l \in\{1,\cdots,\Psi\}$}

\FOR{$n\in\{1,\cdots,\mathcal{N}\}$}
\STATE $\triangleright$ \textit{Update dynamic weights ($\star$)}
\STATE $\beta_{i,l}^n = \exp\left(-\gamma \cdot \dfrac{\|(\boldsymbol{\tau}_{m,l}^{n-1} - \boldsymbol{\tau}_{i,l})\boldsymbol{\tau}_{i,l}^\top\|_F^2}{\|\boldsymbol{\tau}_{i,l}\|_F^2}\right)$ 
\STATE $\boldsymbol{w}_{i,l}^n = \alpha_{i,l} \cdot \beta_{i,l}^n / \sum_{j=1}^\mathcal{P} (\alpha_{j,l} \cdot \beta_{j,l}^n)$ 
\STATE $\triangleright$ \textit{Weighted loss + projection regularization ($\star$)}        

\STATE $\mathcal{L}_{l}^n = \sum_{i=1}^\mathcal{P} \frac{\boldsymbol{w}_{i,l}^n}{\|\boldsymbol{\tau}_{i,l}\|_F^2} \left\| (\boldsymbol{\tau}_{m,l}^{n-1} - \boldsymbol{\tau}_{i,l})\boldsymbol{\tau}_{i,l}^\top \right\|_F^2$
\STATE ${ \mathcal{R}_l^n = \lambda \cdot \sum_{i=1}^\mathcal{P} \left\| \nabla \mathcal{L}_l^n - \frac{\nabla \mathcal{L}_l^n \cdot \boldsymbol{\tau}_{i,l}^\top}{\|\boldsymbol{\tau}_{i,l}\|_F^2} \boldsymbol{\tau}_{i,l} \right\|_F^2} $
\STATE $\triangleright$ \textit{Regularized update ($\star$)}

\STATE $\boldsymbol{\tau}_{m,l}^n = \boldsymbol{\tau}_{m,l}^{n-1} - \zeta \cdot \left ( \nabla \mathcal{L}_l^n + \nabla_{\boldsymbol{\tau}_{m,l}^{n-1}} \mathcal{R}_l^n \right ) $

\ENDFOR

\ENDFOR
\STATE $\triangleright$ Assemble merged task vectors

\STATE $\boldsymbol{\tau}_{m} = \{ \boldsymbol{\tau}_{m,l} \}_{l=1}^\Psi$, $\boldsymbol{\theta_m} = \boldsymbol{\theta} + \boldsymbol{\tau}_{m}$

\end{algorithmic}

\end{algorithm}

The original merge loss $\mathcal{L}_l$ in Eq.~\ref{eq:wudi_merge} only focuses on the compatibility of the merged vector with each task vector. However, different directions in the parameter space have different impacts on task performance, and a slight change in some parameter directions may significantly affect task accuracy. Therefore, we design the gradient projection regularization $\mathcal{R}_l^n$, which decomposes the gradient $\nabla \mathcal{L}_l^n$ into two components. The parallel component $\frac{\nabla \mathcal{L}_l^n \cdot \boldsymbol{\tau}_{i,l}^\top}{\|\boldsymbol{\tau}_{i,l}\|_F^2} \boldsymbol{\tau}_{i,l}$, which represents the update consistent with the direction of the task vector $\boldsymbol{\tau}_{i,l}\). And the orthogonal component $\nabla \mathcal{L}_l^n - \frac{\nabla \mathcal{L}_l^n \cdot \boldsymbol{\tau}_{i,l}^\top}{\|\boldsymbol{\tau}_{i,l}\|_F^2} \boldsymbol{\tau}_{i,l}$, which represents the update perpendicular to the direction of the task vector (causing task interference). By penalizing the norm of the orthogonal component, the optimization process is forced to proceed along the direction of the task vector, thus reducing the updates that are harmful to specific tasks. The penalty term $\mathcal{R}_l^n$ is defined as follows:
\begin{align}
    \mathcal{R}_l^n = \lambda \cdot \sum_{i=1}^\mathcal{P} \left\| \nabla \mathcal{L}_l^n - \frac{\nabla \mathcal{L}_l^n \cdot \boldsymbol{\tau}_{i,l}^\top}{\|\boldsymbol{\tau}_{i,l}\|_F^2} \boldsymbol{\tau}_{i,l} \right\|_F^2
    \label{con:r}
\end{align}

The algorithm flow is described in detail in Algorithm ~\ref{algorithm1}. The optimization of each linear layer is independent, and the problem can be solved layer by layer in turn by using the gradient descent method for optimization.

%% file: src/4_experiment.tex
\section{Experiments}

\begin{table}[t] 
    \caption{Composition of our aggregated dataset from public sources, with their corresponding sizes.}
\centering
\resizebox{1\columnwidth}{!}{%
\begin{tabular}{c|c|c}
\hline
\textbf{Task} & \textbf{Size} & \textbf{Datasets} \\
\hline
\textbf{Math} & 15K & \thead{Geometry3K~\citep{lu2021geo3k} GeoQA+~\citep{cao2022geoqa+} K12~\citep{meng2025mmeureka} UniGeo~\citep{chen2022unigeo}}\\
\hline
\thead{\textbf{Table} 
\textbf{Chart} \\
\textbf{Document}}& 15K & \thead{TAT-DQA~\citep{zhu2022tatdqa} WTQ~\citep{pasupat2015wtq} TabFact~\citep{chen2019tabfact} \\PlotQA~\citep{methani2020plotqa} TQA~\citep{Kembhavi2017TQA} ChartGalaxy~\citep{li2025chartgalaxy}} \\
\hline
\textbf{OCR} & 10K & \thead{TextVQA ~\citep{singh2019textvqa} ChromeWriting TextOCR~\citep{singh2021textocr} \\
OCR-VQA~\citep{mishra2019ocrvqa} }\\ 

\hline
\textbf{All(Mix)} & 40K & - \\
\hline
\end{tabular}
}
\label{tab:dataset_info}
\end{table}

In this section, we first provide the experiments setup in Sec.~\ref{sec:implementation}, and then present main results in Sec \ref{sec:main_result} that demonstrate the effectiveness of Tiny-R1V. In Sec.~\ref{sec:ablation}, we conduct ablation studies to evaluate the impact of each design in Tiny-R1V. Sec.~\ref{sec:qualitative} provides qualitative results of Tiny-R1V.

\subsection{Experiments Setup}
\label{sec:implementation}

\textbf{Implementation.} In this work, we adopt Qwen2.5-VL-3B as our base model. Model training is implemented using the EasyR1~\citep{zheng2025easyr1} codebase, and the training is executed on 8 NVIDIA A800 (40G) GPUs. For the rollout parameters, we set the number of samples per question to 5 and a probability $p$ of 0.3. Regarding RL-related hyperparameters, we use a global batch size of 128, a rollout batch size of 512, a rollout temperature of 0.7, and a learning rate of 1e-6. For training data, we collect a broader range of domain-specific data, which is divided into math, structured data (table, chart, document) and OCR. The datasets used are summarized in Table ~\ref{tab:dataset_info}.

\textbf{Evaluation Benchmark.}
Our model is evaluated across three key dimensions, multimodal mathematical reasoning, multimodal structured data reasoning, and OCR capabilities. For multimodal mathematical reasoning, we compare detailed performance metrics on the MathVista (MINI)~\citep{lu2023mathvista}, MathVision~\citep{wang2024mathvision}, MathVerse (Vision-Only)~\citep{zhang2024mathverse}, and WeMath (Strict)~\citep{qiao2024wemath} benchmarks. In terms of multimodal structured data reasoning, evaluations are conducted on ChartQA (Test Average)~\citep{masry2022chartqa} and CharXiv (Reasoning Questions)~\citep{wang2024charxiv} benchmarks. For OCR capabilities, the model is assessed using the OCRbench~\citep{liu2024ocrbench} and OCR-Reasoning~\citep{huang2025ocr_reasoning} benchmarks. Additionally, to verify that the Tiny-R1V model retains general capabilities, we also report its performance on the MME~\citep{fu2023mme} and MMStar~\citep{chen2024mmstar} benchmarks.

\subsection{Main Results}
\label{sec:main_result}
To comprehensively evaluate the effectiveness of our proposed Tiny-R1V, we conduct extensive comparisons across 10 widely used and challenging benchmarks, as illustrated in Table~\ref{tab:qwen2_5vl_main}.

\textbf{Individual models.} Without using cold start, we apply the data in Table~\ref{tab:dataset_info} and use LIPO to train math model, structure model and OCR model,  respectively. LIPO yields a substantial enhancement in the reasoning capabilities of  MLLMs. For example, in challenging math benchmarks such as MathVista and MathVerse, Math Model (with LIPO) achieves +4.2\% and +4.6\% improvement, respectively. In the challenging reasoning chart understanding benchmark CharXiv, structure Model (with LIPO) obtains +3.2\% improvement. On the OCR-Reasoning task that examines both perception and reasoning capabilities, the three models (with LIPO) improve by +4.4\%, +2.3\%, and +1.1\%, respectively. It is worth noting that although we do not specifically train a model for general ability, we also achieve improvement on the general ability benchmarks MME and MMStar, which shows the generalization ability of LIPO in enhancing reasoning ability across different tasks.


\textbf{Comparison with other merging methods.} We further compare Tiny-R1V with representative model merging methods. The results show that Tiny-R1V outperforms all the compared merging methods across the board, achieving the highest average score of 51.6, which is +0.8\% higher than the second-best method namely WUDI Merging, and +4.9\%  higher than the baseline Task Arithmetic. This constitutes a notable improvement, signifying that the model achieves a distinct average enhancement across 10 benchmarks while fully preserving its various reasoning capabilities. This indicates that our approach effectively merges specialized capabilities, and well alleviates the problem of unbalanced performance across different tasks that exists in other merging methods.

\textbf{Comparison with other models.} We compare Tiny-R1V with other general MLLMs. With less training data, our Tiny-R1V outperforms models such as InternVL2.5-2B and VITA-1.5-8B, with an average performance improvement of +6.7\%. It is worth noting that in addition to having reasoning ability, Tiny-R1V also demonstrates stronger generalization capabilities in different tasks.

\begin{table}[htbp] 
\centering 

\renewcommand{\arraystretch}{1.1}
\caption{Accuracy and average reasoning response token number of models and methods on different Mathematica benchmarks. Models trained via LIPO (ours) are denoted by the bolded items in the table. For a fair comparison, the models labeled as $^{\star}$ are trained using the same experimental data from Table ~\ref{tab:dataset_info}. (Acc $\uparrow$, Avg. Token Length $\downarrow$)} 
\resizebox{1\columnwidth}{!}{ 
\begin{tabular}{l|c|cc|cc}
\toprule
\textbf{Model} & \textbf{Method} & \multicolumn{2}{c|}{\textbf{MathVista}} & \multicolumn{2}{c}{\textbf{MathVision}} \\
\cline{3-6} 
& & \textbf{Acc$\uparrow$} & \textbf{Token$\downarrow$} & \textbf{Acc $\uparrow$} & \textbf{Token$\downarrow$} \\
\hline
Qwen2.5-VL-3B-Inst. & - & 62.3 & 87.5 & 21.2 & 466.5 \\

Qwen2.5-VL-3B-Inst.$^{\star}$ & GRPO & 66.1 & 138.1 & 23.0 & 440.1 \\

R1-VL-2B~\citep{zhang2025r1vl}  & GRPO & 52.1 & 158.8 & 17.1 & 383.2 \\

VLM-R1-3B~\citep{shen2025vlmr1} & GRPO & 62.7 & 147.3 & 21.9 & 349.7 \\

\rowcolor{gray!15}\textbf{Qwen2.5-VL-3B-Inst.}$^{\star}$ & LIPO & 66.5 & 83.0  & 23.3 & 114.9 \\

\rowcolor{gray!15}\textbf{Tiny-R1V-3B (+AMM)}$^{\star}$ & LIPO & 65.5 & 84.5 & 23.7 & 120.2 \\
\hline
\multicolumn{6}{c}{\textit{Scaling to Larger Models}} \\
\hline
Qwen2.5-VL-7B-Inst. & - &68.2 &208.2  &25.1 & 522.2\\
Qwen2.5-VL-7B-Inst.$^{\star}$  & GRPO & 72.8  & 308.1  & 29.4 & 477.8 \\
R1-Onevision-7B~\citep{yang2025r1onevision}  & GRPO & 64.1 &254.6 &29.9&  463.4 \\
Vision-R1-7B~\citep{huang2025visionr1} & GRPO & 73.5 & 410.2 &29.4 & 830.9 \\
\rowcolor{gray!15} \textbf{Qwen2.5-VL-7B-Inst.}$^{\star}$ & LIPO & 72.2 & 121.6 & 29.0 & 147.5\\

\bottomrule
\end{tabular}
}
\label{tab:token_length}
\end{table}

\label{sec:qualitative}
\begin{figure*}[ht]
\centering
\includegraphics[width=0.94\textwidth]{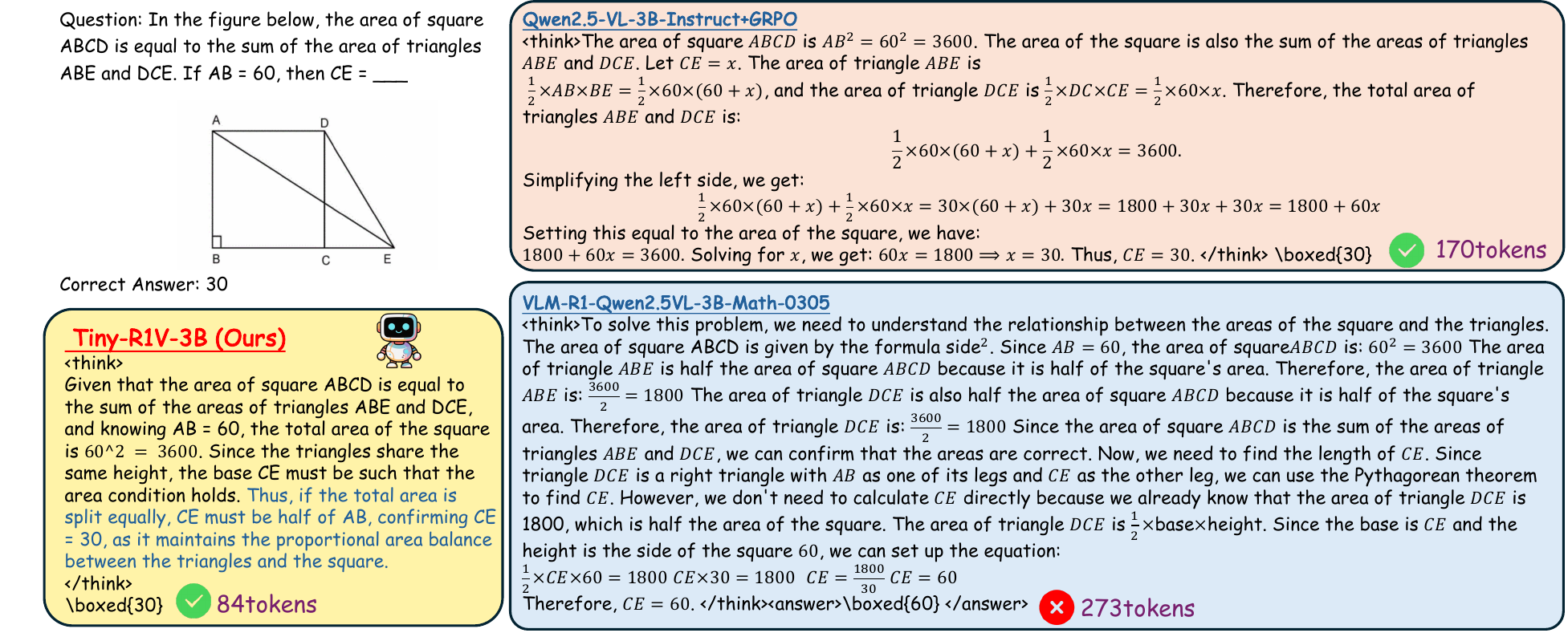} 
\caption{Qualitative Results of reasoning capability of Qwen2.5-VL-3B-Instruct~\citep{bai2025qwen2_5_vl} with GRPO, VLM-R1-Qwen2.5VL-3B-Math-0305 (official open-source weights)~\citep{shen2025vlmr1} and Tiny-R1V-3B (Ours) on mathematical problems.}
\label{fig:qua}
\end{figure*}

\subsection{Response length discussion.}\label{sec:lenght_dis}

The following analysis takes mathematical models as an example to illustrate the effectiveness of the LIPO method in reducing inference tokens. Table~\ref{tab:token_length} clearly demonstrates the remarkable balance between the accuracy and response efficiency achieved by our LIPO-based models. For a fair comparison, the models labeled as $^{\star}$ are trained using the same experimental data from Table ~\ref{tab:dataset_info}. The Qwen2.5-VL-3B model with GRPO consumes an average of 138.1 tokens to reach 66.1\% accuracy on MathVista, whereas our LIPO-enabled Qwen2.5-VL-3B-Instruct and Tiny-R1V-3B models deliver superior or comparable accuracy (66.5\% and 65.5\% on MathVista, respectively) with significantly fewer tokens (83.0 and 84.5 tokens). 

This trend is even more pronounced on MathVision, while baseline models using GRPO need 349.7$\sim$440.1 tokens to reach 21\% $\sim$ 23\% accuracy, our LIPO models hit 23.7\% accuracy with only 114.9$\sim$120.2 tokens with a token cost of less than one-third of the former. Compared with existing R1 mathematical reasoning models, R1-VL-2B~\citep{zhang2025r1vl} and VLM-R1-3B~\citep{shen2025vlmr1}, our models achieve better reasoning performance using only one-third of the token count. Figure~\ref{fig:firstpage}(b) presents the average response length of GRPO~\citep{guo2025deepseekr1}, DAPO~\citep{yu2025dapo}, and our proposed LIPO during the RL training process on the training set. As observed from the figure, the number of tokens used by LIPO gradually decreases throughout the training process, and it achieves a significantly lower token count compared to both GRPO and DAPO methods. This indicates that LIPO effectively optimizes reasoning efficiency by focusing on critical logical steps rather than redundant processes, shaking the inherent notion that higher accuracy necessitates longer responses. 

To further evaluate the effectiveness of our proposed LIPO, we extend the analysis to larger-scale models. As shown in Table~\ref{tab:token_length}, the models trained with our LIPO approach maintain competitive accuracy while significantly reducing the average reasoning token length compared with GRPO-based baseline and other opensource GRPO-based methods. For instance, on the MathVision benchmark, Qwen2.5-VL-7B-Instruct with LIPO achieves a similar accuracy of 29.0 compared to 29.4 from Qwen2.5-VL-7B-Instruct with GRPO and Vision-R1-7B~\citep{huang2025visionr1}, yet reduces the average token length from 463.4$\sim$830.9 to 147.5. This demonstrates that the proposed method not only scales effectively with model size but also enhances reasoning efficiency without compromising performance.


\subsection{Ablation Study}\label{sec:ablation}
\textbf{Ablation Study of AMM.} As shown in Table~\ref{tab:AMM}, we conduct ablation studies to test the contribution of various designs in AMM, including the design of  $\alpha_{i,l}$, $\beta_{i, l}^{n}$ parameters and gradient projection regularization. Compared with the Wudi Model Merging baseline, $\alpha_{i,l}$, $\beta_{i, l}^{n}$ parameters can improve the performance by 1.1\%. In addition, incorporating the regularization penalty,  yields a
performance boost of +0.3\%. Finally, the AMM model merging achieves the best score of 65.5\% on MathVista, reflecting the effectiveness of adaptively adjusting each task vector.

\begin{table}[htbp] 
\caption{Ablation study of Adaptive Model Merging. We study the impact of $\alpha_{i,l},\beta_{i, l}^{n},\mathcal{R}_l$ in AMM. } 
\resizebox{1\columnwidth}{!}{ 
\begin{tabular}{l|ccc}
\toprule
\textbf{Model}  & \textbf{MathVista}  & \textbf{ChartQA} & \textbf{OCRBench}\\
\hline
Qwen2.5-VL-3B-Instruct & 62.3& 84.0& 79.7 \\
Math Model with LIPO &66.5 &83.9 & 73.5 \\
Structure Model with LIPO &61.9 &85.8 & 77.6 \\
OCR Model with LIPO  & 57.9  &83.5  & 84.6 \\ 
Wudi Model Merging  &63.5 & 85.0& 81.5 \\
\hline
+ $\alpha_{i,l}$ &64.9 &85.2 & 81.7 \\ 
+ $\beta_{i, l}^{n}$ &65.2& 85.3& 82.3 \\ 
+ $\mathcal{R}_l$ \& Regularized update & 65.5&85.3 &82.5  \\

\bottomrule
\end{tabular}
}
\label{tab:AMM}
\end{table}

\begin{table}[htbp]
\centering 
\label{tab:params_combined}

\begin{minipage}[t]{0.48\columnwidth} 
\centering
\caption{The impact of $\eta,L_T$ in LIPO.} 
\resizebox{\textwidth}{!}{ 
\begin{tabular}{l|cc}
\toprule
\textbf{$L_T$}  & \textbf{MathVista}  & \textbf{MathVision} \\
\hline
100 & 66.1 & 22.9\\
\textbf{120} & \textbf{66.5} & \textbf{23.3}\\
140 & 66.2 & 23.1 \\
160 & 66.1 & 23.0 \\
\bottomrule
\end{tabular}
}
\label{tab:LT}
\end{minipage}
\hfill 
\begin{minipage}[t]{0.48\columnwidth} 
\centering
\caption{The impact of the threshold $\eta$ in LIPO.} 
\resizebox{\textwidth}{!}{ 
\begin{tabular}{l|cc}
\toprule
\textbf{$\eta$}  & \textbf{MathVista}  & \textbf{MathVision} \\
\hline
0.1 & 66.2& 23.1\\
\textbf{0.2} &  \textbf{66.5} & \textbf{23.3} \\
0.3 & 66.1 &22.8\\
0.4 & 66.0& 22.8\\
\bottomrule
\end{tabular}
}
\label{tab:theta}
\end{minipage}
\end{table}

\textbf{Hyperparameter studies of LIPO.}
To further evaluate the robustness of LIPO, we conduct extensive hyperparameter studies focusing on two critical parameters, the length threshold $  L_T  $ and the reward threshold $  \eta  $. Table~\ref{tab:LT} investigates the impact of $  L_T  $ on model performance across the MathVista and MathVision datasets. Performance peaks at $  L_T = 120  $ with 66.5\% and 23.3\% on the two datasets, respectively, before gradually declining as $  L_T  $ increases further.This trend indicates that $L_T=120$ is sufficient to achieve best mathematical reasoning performance, and it also proves that LIPO is not sensitive to length restrictions and is effective within a suitable range. Table~\ref{tab:theta} analyzes the influence of the confidence threshold $  \eta  $. This shows that a moderate threshold $\eta = 0.2$ can effectively filter responses with similar rewards.

\subsection{Qualitative Results}

Figure~\ref{fig:qua} demonstrates that Tiny-R1V significantly boosts the model's reasoning capabilities when tackling complex mathematical problems. In this example, the model accurately parses the question and obtains the correct answer, showing strong performance in symbolic reasoning and function analysis. Meanwhile, Tiny-R1V uses fewer tokens and simpler thinking methods to obtain the correct results. This proves the superiority of Tiny-R1V in performing lightweight reasoning when handling complex reasoning tasks.

\section{Conclusion}
In this paper, we propose Tiny-R1V, a novel lightweight 3B model that unifies multi-task, multi-modal reasoning capabilities while achieving fewer tokens and higher accuracy. Firstly, Tiny-R1V uses LIPO to adjust the proportion of in-group advantage according to the length of the response, and trains specialized reasoning models with shorter responses. Then, it uses AMM to adaptively adjust the weights of task vectors in each model to integrate multiple specialized models into a unified architecture. We conduct extensive experiments and ablation studies, and the result demonstrate the superiority of our proposed Tiny-R1V on various reasoning benchmarks.

%% file: src/appendix.tex
\appendix
\section*{Appendix}

\section{The Use of Large Language Models (LLMs)}

In this paper, large language model (LLM) ~\citep{hurst2024gpt4o} is only utilized to assist with the refinement of the English language. No LLMs are employed to generate new innovative ideas, and the research process is conducted by the researchers (humans). 

\section{Benchmarks}
Our model is evaluated across four key categories, multimodal mathematical reasoning, multimodal structured data reasoning, OCR capabilities and general capabilities. 

\subsection{Mathematics Benchmarks}
\begin{itemize}
    \item \textbf{MathVista}~\citep{lu2023mathvista} encompasses 6,141 questions spanning diverse domains including arithmetic, geometry, algebra, and statistics.
     \item \textbf{MathVision}~\citep{wang2024mathvision} is a curated collection of 3,040 high-quality mathematical problems derived from real-world mathematics competitions.
    \item \textbf{MathVerse}~\citep{zhang2024mathverse}  comprises 2,612 multimodal mathematical problems, accompanied by 15,672 manually annotated test samples. These samples are categorized into 3 primary question types and 12 subcategories, such as plane geometry, solid geometry, and functions.
    \item \textbf{WeMath}~\citep{qiao2024wemath} comprises 6,500 visual mathematical problems, covering 67 hierarchical knowledge concepts and five levels of knowledge granularity. Complex problems are decomposed into a novel four-dimensional metric to hierarchically evaluate the inherent issues in the reasoning process of MLLMs.
\end{itemize}

\subsection{Multimodal Structured Data Reasoning Benchmarks}

\begin{itemize}
    \item \textbf{ChartQA }~\citep{masry2022chartqa} is a large-scale benchmark encompassing 9.6K human-authored questions alongside 23.1K questions generated from human-written chart summaries. 
    \item \textbf{CharXiv}~\citep{wang2024charxiv} involves 2,323 natural, challenging, and diverse charts sourced from scientific papers. It includes both descriptive questions that assess basic chart elements and reasoning questions that require synthesizing information from complex visual elements within the charts.
\end{itemize}

\subsection{OCR Benchmarks}

\begin{itemize}
    \item \textbf{OCRbench}~\citep{liu2024ocrbench} comprises 1,000 question-answer pairs, with all answers undergoing manual verification and correction, designed to evaluate the OCR capabilities of MLLMs. It encompasses five components: text recognition, scene text-centered VQA, document-oriented VQA, key information extraction, and handwritten mathematical expression recognition. 
    \item \textbf{OCR-Reasoning} ~\citep{huang2025ocr_reasoning} consists of 1,069 manually annotated examples, covering 6 core reasoning capabilities and 18 practical reasoning tasks within text-rich visual scenarios. It is specifically designed to systematically evaluate the performance of MLLMs in reasoning tasks involving text-rich images.
\end{itemize}

\subsection{General Benchmarks}
\begin{itemize}
    \item \textbf{MME}~\citep{fu2023mme} encompasses 14 subtasks, which are designed to measure the perceptual and cognitive capabilities of MLLMs.
    \item \textbf{MMStar}~\citep{chen2024mmstar}  consists of 1,500 challenging samples meticulously curated by humans, encompassing 6 core functionalities and 18 detailed tasks, aiming to evaluate the multi-modal capacities of MLLMs with a carefully balanced and purified selection of samples.
\end{itemize}

\section{Prompts}
\begin{tcolorbox}[title = {Prompt Template used for Length-Informed Relative Policy Optimization (LIPO).}, breakable]
\lstset{
  breaklines=true,
  breakatwhitespace=false,
  columns=flexible,
  breakindent=0pt  
} 
\begin{lstlisting}[numbers=none]
You FIRST think about the reasoning process as an internal monologue and then provide the final answer. The reasoning process MUST BE enclosed within </think> </think> tags. The final answer MUST BE put in \boxed{}.
\end{lstlisting}
\end{tcolorbox}

\section{Inference Time }
We compare the inference time across different benchmarks on eight A800 40G GPUs. As illustrated in Table \ref{tab:time}, Tiny-R1V-3B (with LIPO) exhibits notable inference speed superiority over Qwen2.5-VL-3B-Instruct (with GRPO) across all evaluated benchmarks. Tiny-R1V employs fewer tokens during the inference process, resulting in significantly faster inference on each benchmark.

\begin{table}[htbp] 
\caption{Inference time on different benchmarks} 
\center
\resizebox{\columnwidth}{!}{ 
\begin{tabular}{l|ccc}

\toprule
\textbf{Model}  & \textbf{MathVista}  & \textbf{ChartQA} & \textbf{OCRBench}\\
\hline
Qwen2.5-VL-3B-Instruct(GRPO) & 1788.4s & 977.2s & 788.4s  \\
Tiny-R1V-3B(LIPO,Ours) & 623.4s  & 455.4s & 318.6s  \\

\bottomrule
\end{tabular}
}
\label{tab:time}
\end{table}

\section{Mathematics Reasoning Capabilities Case Studies} 

This section demonstrates the significant enhancement in reasoning capabilities of Tiny-R1V when addressing complex mathematical reasoning tasks. In these case studies, Tiny-R1V-3B accurately parses problems and derives correct answers, showcasing its robust proficiency in symbolic reasoning, geometric comprehension, and mathematical computation. Meanwhile, Tiny-R1V-3B achieves correct results using fewer tokens and simpler reasoning approaches, which demonstrate the superiority of its lightweight reasoning in handling complex reasoning tasks. Specifically, mathematical reasoning tasks require rigorous chain-of-mathematical-thought reasoning. Existing methods (such as Qwen2.5-VL-3B-Instruct~\citep{bai2025qwen2_5_vl}, Qwen2.5VL-3B-GRPO~\citep{shao2024deepseekmath}, VLM-R1-Qwen2.5VL-3B-Math-0305~\citep{shen2025vlmr1} ) tend to employ multi-step complex computations and numerous equation transformation operations when tackling complex reasoning tasks. In contrast, our Tiny-R1V-3B can precisely leverage existing mathematical theorems to simplify the mathematical derivation process (see Case 1 and Case 2 for details). This simplified derivation process substantially reduces the token consumption during reasoning while ensuring the correctness of the reasoning process.
\begin{tcolorbox}[
title = {\textbf{Mathematics Reasoning Capabilities Case1}},
colframe = red!75!black, 
colbacktitle = red!25,   
coltitle = black,         
breakable] 
\includegraphics[width=0.3\textwidth]{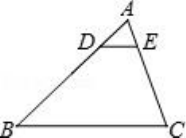}

\textbf{Question:} As shown in the figure, $D$ and $E$ are the points on the edges $AB$ and $AC$ of $\triangle ABC$, $DE \parallel BC$, if $AD:DB=1.0:3.0$, $AE=2.0$, then the length of $AC$ is $(\quad)$.

\tcblower

\textbf{Qwen2.5-VL-3B-Instruct Response}: To solve for the length of \( AC \), we start by analyzing the given information and the geometric relationships in the triangle.

1. \textbf{Identify the coordinates of points \( A \), \( B \), and \( C \):}

\begin{itemize}
    \item Let \( A = (0, 4) \), \( B = (0, 0) \), and \( C = (4, 0) \).
    \item Since \( AD:DB = 1:3 \), point \( D \) divides \( AB \) in the ratio 1:3. Therefore, the coordinates of \( D \) are \( \left(0, \frac{4 \cdot 1}{1+3}\right) = (0, 1) \).
   \item Since \( DE \parallel BC \), point \( E \) divides \( AC \) in the same ratio as \( D \) divides \( AB \), which is 1:3. Therefore, the coordinates of \( E \) are \( \left(\frac{4 \cdot 1}{1+3}, 0\right) = (1, 0) \).
\end{itemize}

2. \textbf{Calculate the length of \( AC \):}
\begin{itemize}
    \item The length of \( AC \) is the distance from \( A = (0, 4) \) to \( C = (4, 0) \). Using the distance formula:
\begin{align}
AC &= \sqrt{(4-0)^2 + (0-4)^2} \\
   &= \sqrt{4^2 + (-4)^2} \\
   &= \sqrt{16 + 16} \\
   &= \sqrt{32} \\
   &= 4\sqrt{2}
\end{align}

\end{itemize}

Thus, the length of \( AC \) is \(\boxed{4\sqrt{2}}\).

{\quad \quad \large \textbf{\textcolor{Red}{\ [Incorrect]}}} 
\tcbline
\textbf{Qwen2.5VL-3B-GRPO Response}:
\texttt{<think>}
Since \(DE \parallel BC\), triangles \(ADE\) and \(ABC\) are similar by the AA similarity criterion. Therefore, the ratios of corresponding sides are equal. Given \(AD:DB = 1:3\), we can infer that \(AD = \frac{1}{4}AB\) and \(DB = \frac{3}{4}AB\). Since \(AB = AD + DB = 4\) cm, we have \(AD = 1\) cm and \(DB = 3\) cm. Because \(DE \parallel BC\), the ratio \(AD:AB = AE:AC\). Given \(AE = 2\) cm, we can set up the proportion:

\[
\frac{AD}{AB} = \frac{AE}{AC} \implies \frac{1}{4} = \frac{2}{AC} \implies AC = 8 \text{ cm}
\]

Thus, the length of \(AC\) is 8 cm.
\texttt{</think>}
\boxed{8} {\quad \quad \large \textbf{\textcolor{ForestGreen}{\ [Correct]}}}

\tcbline
\textbf{VLM-R1-Qwen2.5VL-3B-Math-0305}: \texttt{<think>}
1. Given that \(AD:DB = 1:3\), we can determine the length of \(AD\) and \(DB\).
2. Since \(AD + DB = AB\) and \(AB = 4\), we can write:
   \[
   AD + DB = 4
   \]
3. Given \(AD:DB = 1:3\), we can express \(AD\) and \(DB\) in terms of a common variable \(x\):
   \[
   AD = x \quad \text{and} \quad DB = 3x
   \]
4. Substituting these into the equation \(AD + DB = 4\):
   \[
   x + 3x = 4
   \]
5. Solving for \(x\):
   \[
   4x = 4 \implies x = 1
   \]
6. Therefore, \(AD = 1\) and \(DB = 3\).

Since \(DE \parallel BC\), \(DE\) is parallel to \(BC\) and \(DE\) is half the length of \(BC\) because \(AD:DB = 1:3\).

7. Given \(AE = 2\), we can determine the length of \(AC\):
   \[
   AC = AE + EC
   \]
8. Since \(DE \parallel BC\), \(EC = DB = 3\).
9. Therefore:
   \[
   AC = AE + EC = 2 + 3 = 5
   \]

\texttt{</think>}
\texttt{<answer>}5\texttt{</answer>}  

{\quad \quad \large \textbf{\textcolor{Red}{\ [Incorrect]}}} 

\tcbline

\textbf{\large Tiny-R1V-3B Response (Ours):} \texttt{<think>} Given that $DE \parallel BC$, it implies that $\triangle ADE \sim \triangle ABC$ by the Basic Proportionality Theorem (or Thales' theorem). Therefore, the ratios of corresponding sides are equal. Since $AD:DB=1:3$, $AD:AB=1:4$. Thus, $AE:AC=1:4$. Given $AE=2$, we can find $AC$ as $AC=4 \times AE=4 \times 2=8$. \texttt{</think>}

Answer: $\boxed{8}$ {\quad \quad \large \textbf{\textcolor{ForestGreen}{\ [Correct]}}}

\end{tcolorbox}

\begin{tcolorbox}[
title = {\textbf{Mathematics Reasoning Capabilities Case2}},
colframe = red!75!black, 
colbacktitle = red!25,   
coltitle = black,         
breakable] 
\includegraphics[width=0.4\textwidth]{case_fig/math_fig2.pdf}

\textbf{Question:} As shown in the figure, $PA$ and $PB$ are tangents of $\odot O$, with tangent points $A$ and $B$, $AC$ is the diameter of $\odot O$, given that $\angle P=50.0$, then the size of $\angle ACB$ is $(\quad)$.

\tcblower
\textbf{Qwen2.5-VL-3B-Instruct Response}: To solve for the size of \(\angle ACB\), we start by analyzing the given information and the geometric properties of the circle and the tangents.

1. \textbf{Identify the properties of the tangents and the circle:}
   \begin{itemize}
       \item  \(PA\) and \(PB\) are tangents to the circle \(\odot O\) at points \(A\) and \(B\) respectively.
       \item  \(AC\) is the diameter of the circle.
       \item  \(\angle P = 50^\circ\).
   \end{itemize}

2. \textbf{Use the property of tangents:}
   \begin{itemize}
       \item  The tangents drawn from an external point to a circle are equal in length. Therefore, \(PA = PB\).
       \item  The line segment \(OA\) is perpendicular to the tangent \(PA\) at the point of tangency \(A\), so \(\angle OAP = 90^\circ\).
       \item  Similarly, \(\angle OBP = 90^\circ\).
   \end{itemize}

3. \textbf{Determine the angles in \(\triangle OAP\) and \(\triangle OBP\):}
   \begin{itemize}
       \item  Since \(\angle OAP = 90^\circ\) and \(\angle OBP = 90^\circ\), the sum of the angles in \(\triangle OAP\) and \(\triangle OBP\) is \(180^\circ\).
       \item  Therefore, \(\angle AOP = 180^\circ - \angle P = 180^\circ - 50^\circ = 130^\circ\).
   \end{itemize}

4. \textbf{Use the fact that \(AC\) is the diameter:}
   \begin{itemize}
       \item  Since \(AC\) is the diameter, \(\angle ABC = 90^\circ\) (angle in a semicircle).
       \item  Now, consider \(\triangle OAC\). Since \(OA = OB\) (both are radii of the circle), \(\triangle OAC\) is isosceles with \(\angle OAC = \angle OBA\).
       \item  The sum of the angles in \(\triangle OAC\) is \(180^\circ\). Since \(\angle OAC = \angle OBA\) and \(\angle OAP = 90^\circ\), we have \(\angle OAC = \angle OBA = \frac{180^\circ - 90^\circ}{2} = 45^\circ\).
       \item  Therefore, \(\angle ACB = \angle OAC + \angle OBA = 45^\circ + 45^\circ = 90^\circ - \angle P/2 = 90^\circ - 25^\circ = 65^\circ\).
   \end{itemize}

Thus, the size of \(\angle ACB\) is \(\boxed{65^\circ}\). {\quad \quad \large \textbf{\textcolor{ForestGreen}{\ [Correct]}}}

\tcbline
\textbf{Qwen2.5VL-3B-GRPO Response}:
\texttt{<think>}

Since \(PA\) and \(PB\) are tangents to \(\odot O\) from point \(P\), \(PA = PB\). Therefore, \(\triangle PAB\) is isosceles with \(\angle P = 50^\circ\). The sum of the angles in a triangle is \(180^\circ\), so we have:
\[
\angle PAB = \angle PBA = \frac{180^\circ - 50^\circ}{2} = 65^\circ.
\]
Since \(AC\) is a diameter, \(\angle ABC = 90^\circ\) (angle in a semicircle). Now, \(\angle ACB\) can be found by noting that \(\angle ACB = 180^\circ - \angle CAB - \angle ABC\). Since \(\angle CAB = \angle PAB = 65^\circ\), we have:
\[
\angle ACB = 180^\circ - 65^\circ - 90^\circ = 25^\circ.
\]
\texttt{</think>}
\boxed{25^\circ} {\quad \quad \large \textbf{\textcolor{Red}{\ [Incorrect]}}}
\tcbline

\textbf{VLM-R1-Qwen2.5VL-3B-Math-0305}: 
\texttt{<think>}
To solve for the size of \(\angle ACB\), we start by noting that \(PA\) and \(PB\) are tangents to the circle \(\odot O\) at points \(A\) and \(B\) respectively. This means that \(OA\) and \(OB\) are perpendicular to \(PA\) and \(PB\) respectively. Therefore, \(\angle OAP = 90^\circ\) and \(\angle OBP = 90^\circ\).

Since \(AC\) is the diameter of the circle, \(\angle ABC = 90^\circ\) because any angle inscribed in a semicircle is a right angle. Now, we need to find \(\angle ACB\). We know that \(\angle P = 50^\circ\), and since \(PA\) and \(PB\) are tangents, \(\angle OAP = 90^\circ\) and \(\angle OBP = 90^\circ\). The angle \(\angle APB\) is the external angle to \(\triangle OAP\) and \(\triangle OBP\), and it is equal to the sum of the opposite internal angles, which are \(\angle OAP\) and \(\angle OBP\). Therefore, \(\angle APB = 180^\circ - \angle OAP - \angle OBP = 180^\circ - 90^\circ - 90^\circ = 180^\circ - 180^\circ + 50^\circ = 50^\circ\).

Now, we can use the fact that the sum of the angles in \(\triangle APB\) is \(180^\circ\). So, \(\angle BAP + \angle ABP + \angle APB = 180^\circ\). Since \(\angle BAP = \angle OAP = 90^\circ\) and \(\angle ABP = \angle OBP = 90^\circ\), we have \(\angle BAP + \angle ABP = 180^\circ - \angle APB = 180^\circ - 50^\circ = 130^\circ\). Therefore, \(\angle ACB = 180^\circ - \angle ABC - \angle BAC = 180^\circ - 90^\circ - 40^\circ = 50^\circ\).

Thus, the size of \(\angle ACB\) is \(\boxed{40^\circ}\).
\texttt{</think>}

\texttt{<answer>} $40^\circ$ \texttt{</answer>} {\quad \quad \large \textbf{\textcolor{Red}{\ [Incorrect]}}}

\tcbline

\textbf{\large Tiny-R1V-3B Response (Ours):} \texttt{<think>} Since $PA$ and $PB$ are tangents, $\angle OAP$ and $\angle OBP$ are both $90^\circ$. Thus, $\angle AOB$ is supplementary to $\angle P$, making it $130^\circ$. Since $AC$ is the diameter, $\angle ACB$ is an inscribed angle subtending the same arc, thus $\angle ACB$ is half of the central angle, which is $65^\circ$. Therefore, $\angle ACB$ is half of the remaining angle at the circumference. \texttt{</think>}

Answer: $\boxed{65^\circ} $ {\quad \quad \large \textbf{\textcolor{ForestGreen}{\ [Correct]}}}

\end{tcolorbox}

\begin{tcolorbox}[
title = {\textbf{Mathematics Reasoning Capabilities Case3}},
colframe = red!75!black, 
colbacktitle = red!25,   
coltitle = black,         
breakable] 
\includegraphics[width=0.4\textwidth]{case_fig/math_fig6.pdf}

\textbf{Question:} As shown in the figure, in $\odot O$ with a radius of 5.0, the length of the chord $AB$ is 8.0, then the distance from the center $O$ to the chord $AB$ is $(\quad)$.

\tcblower
\textbf{Qwen2.5-VL-3B-Instruct Response}: To find the distance from the center \(O\) to the chord \(AB\) in the circle, we can use the Pythagorean theorem. Here's the step-by-step solution:

1. \textbf{Identify the given information:}
   \begin{itemize}
       \item  The radius of the circle \(OA = OB = 5.0\).
       \item  The length of the chord \(AB = 8.0\).
   \end{itemize}

2. \textbf{Draw the perpendicular from the center \(O\) to the chord \(AB\):}
   \begin{itemize}
       \item  Let the perpendicular from \(O\) to \(AB\) intersect \(AB\) at point \(M\). Since \(OM\) is perpendicular to \(AB\), \(M\) is the midpoint of \(AB\).
       \item  Therefore, \(AM = MB = \frac{AB}{2} = \frac{8.0}{2} = 4.0\).
   \end{itemize}

3.\textbf{ Form a right triangle \(OAM\):}
    \begin{itemize}
        \item In the right triangle \(OAM\), \(OA\) is the hypotenuse, \(AM\) is one leg, and \(OM\) is the other leg.
        \item We know \(OA = 5.0\) and \(AM = 4.0\).
        \item We need to find the length of \(OM\).
    \end{itemize}

4. \textbf{Apply the Pythagorean theorem:}
   \[
   OA^2 = OM^2 + AM^2
   \]
   Substituting the known values:
   \[
   5.0^2 = OM^2 + 4.0^2
   \]
   \[
   25.0 = OM^2 + 16.0
   \]
   \[
   OM^2 = 25.0 - 16.0
   \]
   \[
   OM^2 = 9.0
   \]
   \[
   OM = \sqrt{9.0} = 3.0
   \]

Therefore, the distance from the center \(O\) to the chord \(AB\) is \(\boxed{3.0}\). {\quad \quad \large \textbf{\textcolor{ForestGreen}{\ [Correct]}}}

\tcbline
\textbf{Qwen2.5VL-3B-GRPO Response}:
\texttt{<think>}
To find the distance from the center \(O\) to the chord \(AB\), we can use the geometric properties of the circle. Specifically, we can use the fact that the perpendicular from the center of the circle to a chord bisects the chord. Let's denote the distance from the center \(O\) to the chord \(AB\) as \(d\). The perpendicular from \(O\) to \(AB\) will split \(AB\) into two equal segments, each of length \(4\) (since \(AB = 8\)).

We can now consider the right triangle \(OAC\) where \(OC\) is the distance from the center to the chord, \(OA\) is the radius of the circle, and \(AC\) is half the length of the chord \(AB\). By the Pythagorean theorem, we have:
\[
OA^2 = OC^2 + AC^2
\]
Substituting the known values \(OA = 5\) and \(AC = 4\), we get:
\[
5^2 = d^2 + 4^2
\]
\[
25 = d^2 + 16
\]
\[
d^2 = 25 - 16
\]
\[
d^2 = 9
\]
\[
d = \sqrt{9} = 3
\]
So, the distance from the center \(O\) to the chord \(AB\) is \(3\).
\texttt{</think>} \boxed{3} {\quad \quad \large \textbf{\textcolor{ForestGreen}{\ [Correct]}}}

\tcbline
\textbf{VLM-R1-Qwen2.5VL-3B-Math-0305}: \texttt{<think>}
To find the distance from the center \( O \) to the chord \( AB \), we can use the Pythagorean theorem. Here's the step-by-step reasoning:

1. \textbf{Identify the given information:}
   \begin{itemize}
       \item  The radius of the circle \( \odot O \) is 5.0.
       \item  The length of the chord \( AB \) is 8.0.
   \end{itemize}

2. \textbf{Draw the perpendicular from the center \( O \) to the chord \( AB \):}
   \begin{itemize}
       \item  Let the perpendicular from \( O \) to \( AB \) intersect \( AB \) at point \( M \). Since \( OM \) is perpendicular to \( AB \), \( M \) is the midpoint of \( AB \).
       \item  Therefore, \( AM = MB = \frac{AB}{2} = \frac{8.0}{2} = 4.0 \).
   \end{itemize}

3. \textbf{Form a right triangle \( OMA \):}
   \begin{itemize}
       \item  In the right triangle \( OMA \), \( OA \) is the radius of the circle, so \( OA = 5.0 \).
       \item  \( AM \) is half the length of the chord \( AB \), so \( AM = 4.0 \).
       \item  \( OM \) is the distance from the center \( O \) to the chord \( AB \), which we need to find.
   \end{itemize}

4. \textbf{Apply the Pythagorean theorem:}
   \begin{itemize}
       \item  The Pythagorean theorem states that in a right triangle, the square of the hypotenuse (the side opposite the right angle) is equal to the sum of the squares of the other two sides.
       \item  In triangle \( OMA \), \( OA \) is the hypotenuse, \( OM \) is one leg, and \( AM \) is the other leg.
       \item  Therefore, \( OA^2 = OM^2 + AM^2 \).
       \item  Substituting the known values, we get \( 5.0^2 = OM^2 + 4.0^2 \).
       \item  Simplifying, we get \( 25.0 = OM^2 + 16.0 \).
       \item  Solving for \( OM^2 \), we get \( OM^2 = 25.0 - 16.0 = 9.0 \).
       \item  Taking the square root of both sides, we get \( OM = \sqrt{9.0} = 3.0 \).
   \end{itemize}

Therefore, the distance from the center \( O \) to the chord \( AB \) is \( 3.0 \).

\texttt{</think>}
\texttt{<answer>}3.0\texttt{</answer>} {\quad \quad \large \textbf{\textcolor{ForestGreen}{\ [Correct]}}}

\tcbline

\textbf{\large Tiny-R1V-3B Response (Ours):} \texttt{<think>}Given a circle with radius 5 and chord $AB=8$, we can use the right triangle formed by the radius, the perpendicular from the center to the chord, and half of the chord. The perpendicular bisects $AB$ into two segments of 4 each, so we use the Pythagorean theorem: $5^2=4^2+d^2$, where $d$ is the distance from $O$ to $AB$. Thus, $25=16+d^2$, so $d^2=9$ and $d=3$. \texttt{</think>}

Answer: $\boxed{3} $ {\quad \quad \large \textbf{\textcolor{ForestGreen}{\ [Correct]}}}

\end{tcolorbox}

\section{Table, Chart and Document Reasoning Capabilities Case Studies} 
This section demonstrates a significant improvement in the reasoning capabilities of Tiny-R1V when tackling table, chart, and document reasoning tasks. In these case studies, Tiny-R1V-3B can accurately parse questions and derive correct answers. Structured data reasoning tasks require both perception and reasoning capabilities; however, the perception process may consume a large number of redundant tokens. As the cases below, Tiny-R1V-3B has reduced such redundant perception processes, thereby achieving a more efficient perception process and a more accurate reasoning process.

\begin{tcolorbox}[
title = {\textbf{Table, Chart and Document Reasoning Capabilities Capabilities Case1}},
colframe = blue!75!black, 
colbacktitle = blue!25,   
coltitle = black,         
breakable] 
\includegraphics[width=0.7\textwidth]{case_fig/case1-2.png}\\

\textbf{Question:} What is the name of the chart with the highest percentage for its largest category?

\tcblower
\textbf{Qwen2.5-VL-3B-Instruct Response}: The chart with the highest percentage for its largest category is the one on the right, which is labeled "Product Differentiation (Rauch 1999)". {\quad \quad \large \textbf{\textcolor{Red}{\ [Incorrect]}}}
\tcbline

\textbf{Qwen2.5VL-3B-GRPO Response}:
\texttt{<think>} 
Looking at the two pie charts, the left one shows "Commodity-Dependence (UN)" with 81\% for Commodity and 19\% for Non-Commodity. The right one shows "Product Differentiation (Rauch 1999)" with 70\% for World Price, 22\% for Differentiated, and 8\% for Reference Price. The chart with the highest percentage for its largest category is the left one, as it has 81\% for Commodity.
\texttt{</think>}
\boxed{\text{Commodity-Dependence (UN)}} {\quad \quad \large \textbf{\textcolor{ForestGreen}{\ [Correct]}}}

\tcbline
\textbf{\large Tiny-R1V-3B Response (Ours):}\texttt{<think>} The chart on the left shows a clear majority of the category labeled "Commodity" at 81\%, indicating it's the largest segment. The right chart shows a significant portion for "World Price" at 70\%, but the left chart's largest category is more dominant.\texttt{</think>} 

Answer: $\boxed{\text{Commodity-Dependence (UN)}}$ {\quad \quad \large \textbf{\textcolor{ForestGreen}{\ [Correct]}}}
\end{tcolorbox}

\begin{tcolorbox}[
title = {\textbf{Table, Chart and Document Reasoning Capabilities Capabilities Case2}},
colframe = blue!75!black, 
colbacktitle = blue!25,   
coltitle = black,         
breakable] 
\includegraphics[width=0.7\textwidth]{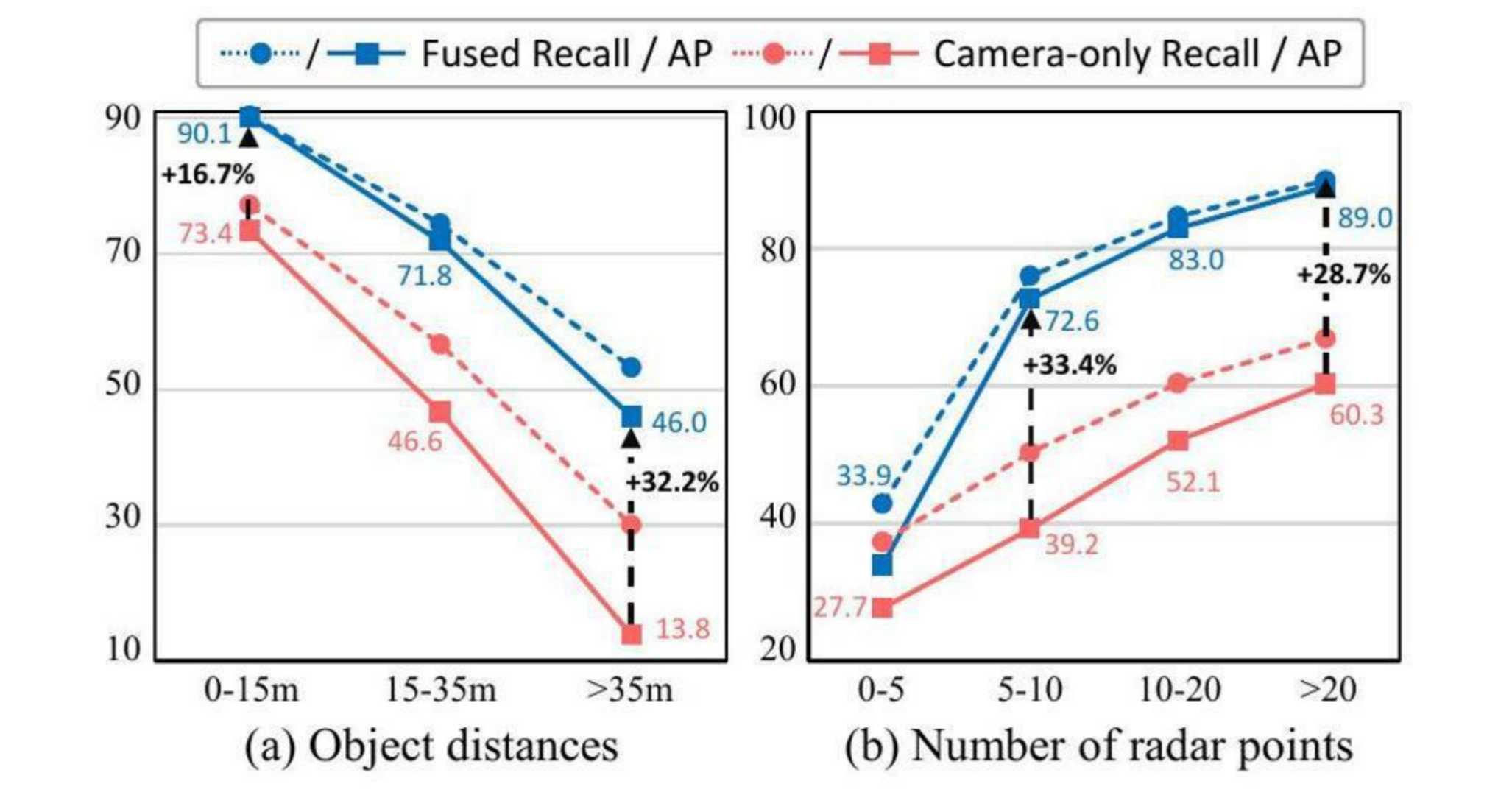}\\

\textbf{Question:} Does increase the number of radar points close the gap between Fused recall and Camera-only recall?

\tcblower

\textbf{Qwen2.5-VL-3B-Instruct Response}: Yes, increasing the number of radar points generally closes the gap between Fused Recall and Camera-only Recall. This can be seen in both graphs:

1. \textbf{Graph (a): Object Distances}
   \begin{itemize}
       \item  The Fused Recall/AP line (solid blue) shows a significant improvement as the object distance increases.
       \item  The Camera-only Recall/AP line (dashed red) shows a corresponding decrease in performance.
       \item  As the object distance increases, the gap between the two lines narrows, indicating that the fused approach performs better.
   \end{itemize}

2.\textbf{ Graph (b): Number of Radar Points}
   \begin{itemize}
       \item  The Fused Recall/AP line (solid blue) shows a significant improvement as the number of radar points increases.
       \item  The Camera-only Recall/AP line (dashed red) shows a corresponding decrease in performance.
       \item  As the number of radar points increases, the gap between the two lines narrows, indicating that the fused approach performs better.
   \end{itemize}

In both cases, the fused approach (solid lines) consistently outperforms the camera-only approach (dashed lines) as the number of radar points or object distances increase. {\quad \quad \large \textbf{\textcolor{ForestGreen}{\ [Correct]}}}
\tcbline
\textbf{Qwen2.5VL-3B-GRPO Response}: \texttt{<think>}
The graph shows two lines: one for "Fused Recall / AP" (solid blue line) and one for "Camera-only Recall / AP" (dotted red line). The solid blue line starts at 90.1\% for 0-15m and decreases to 46.0\% for $>$35m. The dotted red line starts at 73.4\% for 0-15m and decreases to 13.8\% for $>$35m. The gap between the two lines is significant, especially at shorter distances. As the number of radar points increases (from 0-5 to $>$20), the solid blue line (Fused Recall) increases from 33.9\% to 89.0\%, while the dotted red line (Camera-only Recall) increases from 27.7\% to 60.3\%. This indicates that increasing the number of radar points does indeed help close the gap between the Fused Recall and Camera-only Recall. 
\texttt{</think>}
\boxed{\text{Yes}} {\quad \quad \large \textbf{\textcolor{ForestGreen}{\ [Correct]}}}
\tcbline
\textbf{\large Tiny-R1V-3B Response (Ours):} \texttt{<think>} The graphs show that as the number of radar points increases, the Fused Recall/AP line approaches the Camera-only Recall/AP line, indicating a closing gap. This suggests that more radar data points improve the recall and AP metrics, aligning the performance closer to the camera-only results. The percentage increases also support this, showing improvements as the number of points grows.\texttt{</think>}

Answer: $\boxed{\text{Yes}}$ {\quad \quad \large \textbf{\textcolor{ForestGreen}{\ [Correct]}}}
\end{tcolorbox}

\begin{tcolorbox}[
title = {\textbf{Table, Chart and Document Reasoning Capabilities Capabilities Case3}},
colframe = blue!75!black, 
colbacktitle = blue!25,   
coltitle = black,         
breakable] 
\includegraphics[width=\textwidth]{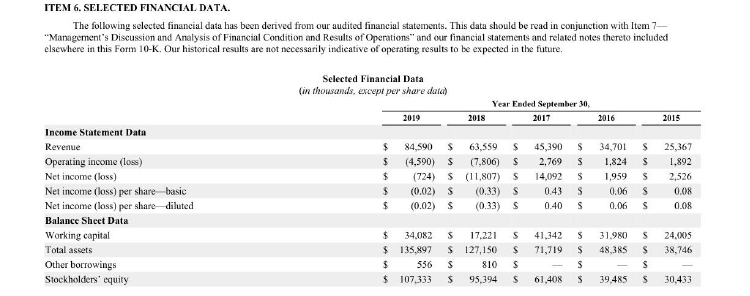}\\

\textbf{Question:} What is the percentage change in working capital from 2018 to 2019?

\tcblower
\textbf{Qwen2.5-VL-3B-Instruct Response}: To calculate the percentage change in working capital from 2018 to 2019, we use the formula:

\[
\text{Percent Change} = \left( \frac{\text{New Value} - \text{Old Value}}{\text{Old Value}} \right) \times 100
\]

Given:
 \begin{itemize}
     \item Working capital in 2019: $\$34,082$
     \item Working capital in 2018: $\$17,221$
 \end{itemize}

Plugging in the values:

\[
\text{Percentage Change} = \left( \frac{34,082 - 17,221}{17,221} \right) \times 100
\]

\[
\text{Percentage Change} = \left( \frac{16,861}{17,221} \right) \times 100
\]

\[
\text{Percentage Change} \approx 97.48\%
\]

So, the percentage change in working capital from 2018 to 2019 is approximately 97.48\%.

{\quad \quad \large \textbf{\textcolor{Red}{\ [Incorrect]}}}
\tcbline
\textbf{Qwen2.5VL-3B-GRPO Response}:
\texttt{<think>}
To calculate the percentage change in working capital from 2018 to 2019:
Percentage change = (New value - Original value) / Original value * 100 \\
Working capital in 2018 = $\$17,221$  \\
Working capital in 2019 = $\$34,082$ \\
Percentage change = ($\$34,082$ - $\$17,221$) / $\$17,221$ * 100\\
Percentage change = $\$16,861$ / $\$17,221$ * 100\\
Percentage change $\approx $ 97.28\%\\
\texttt{</think>}
\boxed{97.28\%} 

{\quad \quad \large \textbf{\textcolor{Red}{\ [Incorrect]}}}
\tcbline 

\textbf{\large Tiny-R1V-3B Response (Ours):} \texttt{<think>} The working capital in 2018 was \$17,221 thousand and in 2019 it was \$34,082 thousand. The change is calculated as (34,082 - 17,221) / 17,221 * 100 = 97.9\%. This indicates an increase of approximately 97.9\% in working capital from 2018 to 2019.\texttt{</think>}

Answer: $  \boxed{97.9\%}  $ {\quad \quad \large \textbf{\textcolor{ForestGreen}{\ [Correct]}}}
\end{tcolorbox}

\begin{tcolorbox}[
title = {\textbf{Table, Chart and Document Reasoning Capabilities Capabilities Case4}},
colframe = blue!75!black, 
colbacktitle = blue!25,   
coltitle = black,         
breakable] 
\includegraphics[width=\textwidth]{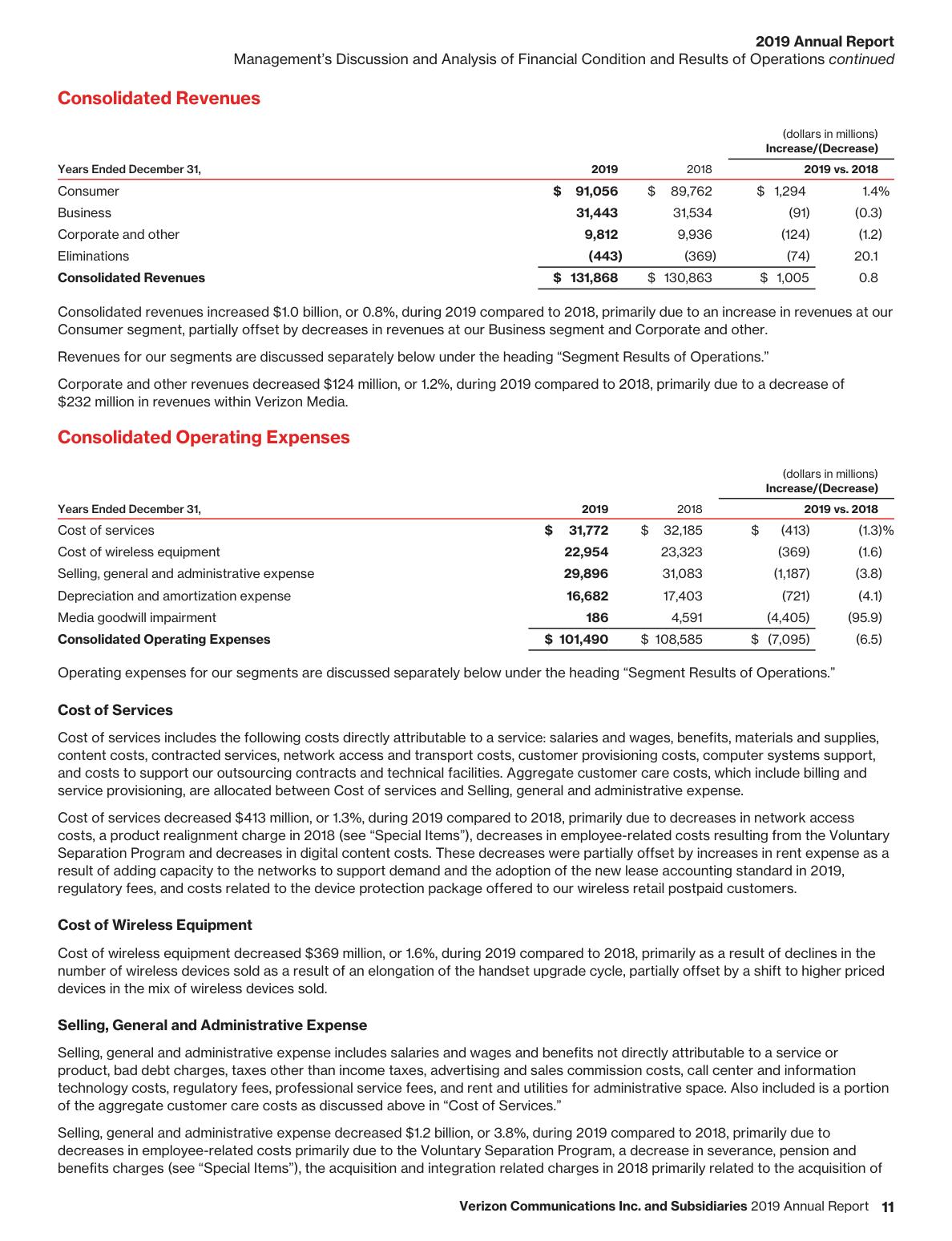}\\

\textbf{Question:} What was the average corporate segment revenue and other segment revenue for 2018 and 2019?
\tcblower

\textbf{Qwen2.5-VL-3B-Instruct Response}: To calculate the average corporate segment revenue and other segment revenue for 2018 and 2019, we need to follow these steps:

1. Identify the revenue for the corporate and other segment for both years.
2. Calculate the average of these two values.

From the table:
\begin{itemize}
    \item  Corporate and other revenue for 2019: $\$9,812$ million
    \item  Corporate and other revenue for 2018: $\$9,936$ million
\end{itemize}

Now, we calculate the average:

\[
\text{Average} = \frac{\text{2019 Revenue} + \text{2018 Revenue}}{2}
\]

\[
\text{Average} = \frac{9,812 + 9,936}{2}
\]

\[
\text{Average} = \frac{19,748}{2}
\]

\[
\text{Average} = 9,874 \text{ million dollars}
\]

So, the average corporate segment revenue and other segment revenue for 2018 and 2019 is $\$9,874$ million. {\quad \quad \large \textbf{\textcolor{ForestGreen}{\ [Correct]}}}

\tcbline
\textbf{Qwen2.5VL-3B-GRPO Response}:
\texttt{<think>}
To find the average corporate and other segment revenue for 2018 and 2019, we need to add the revenues for both years and divide by 2.
For 2019, the corporate and other segment revenue is $\$9,812$ million.
For 2018, the corporate and other segment revenue is $\$9,936$ million.
Average = (9,812 + 9,936) / 2
Average = 19,748 / 2
Average = $\$9,874$ million
Therefore, the average corporate and other segment revenue for 2018 and 2019 is $\$9,874$ million.
\texttt{</think>}
\boxed{9874}  {\quad \quad \large \textbf{\textcolor{ForestGreen}{\ [Correct]}}}

\tcbline

\textbf{\large Tiny-R1V-3B Response (Ours):} \texttt{<think>}
To find the average, I'll add the revenues for both segments in 2018 and 2019, then divide by 2. For Corporate and other, it's (9,936 + 9,812) / 2 = 9874. 
\texttt{</think>}
\boxed{9874} {\quad \quad \large \textbf{\textcolor{ForestGreen}{\ [Correct]}}}
\end{tcolorbox}
\section{OCR Capabilities Case Studies}

This section demonstrates the capability of Tiny-R1V in OCR tasks. In these case studies, Tiny-R1V-3B can accurately recognize image content and derive correct answers, and exhibits accurate performance in tasks such as formula recognition, image-based question answering, and character parsing.

\begin{tcolorbox}[
title = {\textbf{OCR Capabilities Case1}},
colframe = orange!75!black, 
colbacktitle = orange!25,   
coltitle = black,         
breakable] 
\includegraphics[width=0.7\textwidth]{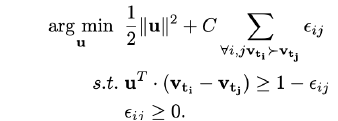}\\

\textbf{Question:} Please write out the expression of the formula in the image using LaTeX format.

\tcblower
\textbf{Tiny-R1V-3B Response:} \texttt{<think>} 
\begin{align}
&\arg\min_{u} \frac{1}{2}\|u\|^2 + C\sum_{\forall i,j \mathbf{v}_{ti} > \mathbf{v}_{tj}} \epsilon_{ij} \\
&\text{s.t. } \mathbf{u}^T \cdot (\mathbf{v}_{ti} - \mathbf{v}_{tj}) \geq 1 - \epsilon_{ij} \\
&\epsilon_{ij} \geq 0.
\end{align}
\texttt{</think>}

Answer: \boxed{
\begin{aligned}
&\arg\min_{u} \frac{1}{2}\|u\|^2 + C\sum_{\forall i,j \mathbf{v}_{ti} > \mathbf{v}_{tj}} \epsilon_{ij} \\
&\text{s.t. } \mathbf{u}^T \cdot (\mathbf{v}_{ti} - \mathbf{v}_{tj}) \geq 1 - \epsilon_{ij} \\
&\epsilon_{ij} \geq 0.
\end{aligned}
} {\quad \quad \large \textbf{\textcolor{ForestGreen}{\ [Correct]}}}
\end{tcolorbox}

\begin{tcolorbox}[
title = {\textbf{OCR Capabilities Case2}},
colframe = orange!75!black, 
colbacktitle = orange!25,   
coltitle = black,         
breakable] 
\includegraphics[width=0.4\textwidth]{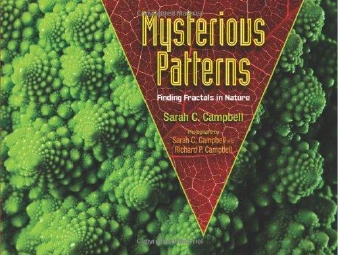}\\

\textbf{Question:} Please write out the expression of the formula in the image using LaTeX format.

\tcblower
\textbf{Tiny-R1V-3B Response:} \texttt{<think>} Sarah C. Campbell \texttt{</think>}

Answer: \boxed{\text{Sarah C. Campbell}} {\quad \quad \large \textbf{\textcolor{ForestGreen}{\ [Correct]}}}
\end{tcolorbox}

\begin{tcolorbox}[
title = {\textbf{OCR Capabilities Case3}},
colframe = orange!75!black, 
colbacktitle = orange!25,   
coltitle = black,         
breakable] 
\includegraphics[width=0.4\textwidth]{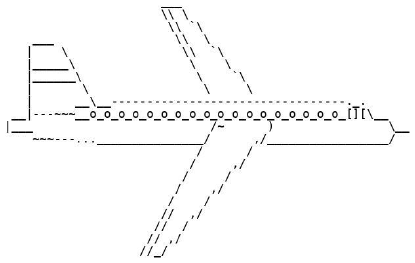}\\

\textbf{Question:} Which of the following options best matches the image? Here's some options: submarine, airplane, ship, chopper.

\tcblower
\textbf{Tiny-R1V-3B Response:} \texttt{<think>} The image depicts a shape that resembles an airplane with wings and a tail. The dashed lines and the overall structure suggest an aircraft. Among the given options, the one that best matches the image is an airplane. \texttt{</think>}

Answer: \boxed{\text{airplane}} {\quad \quad \large \textbf{\textcolor{ForestGreen}{\ [Correct]}}}
\end{tcolorbox}

\begin{tcolorbox}[
title = {\textbf{OCR Capabilities Case4}},
colframe = orange!75!black, 
colbacktitle = orange!25,   
coltitle = black,         
breakable] 
\includegraphics[width=0.4\textwidth]{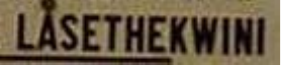}\\

\textbf{Question:} How many times does the character 'e' appear in the picture? Please output the exact number without any additional explanation.

\tcblower
\textbf{Tiny-R1V-3B Response:} \texttt{<think>} I need to count the occurrences of the character 'e' in the given text. I'll go through each character and count how many times 'e' appears. \texttt{</think>}

Answer: \boxed{\text{2}} {\quad \quad \large \textbf{\textcolor{ForestGreen}{\ [Correct]}}}
\end{tcolorbox}

\section{Limitations}

Due to resource constraints, our experiments are limited to 3B-parameter models. Future work will explore the integration of MLLMs with larger parameter sizes, multilingual support, and a broader range of reasoning tasks. By incorporating more advanced chain-of-thought techniques, such as thinking with images, we aim to enhance its reasoning capabilities on more complex visual-logical reasoning tasks. Additionally, we plan to filter out low-quality samples using rules during the preprocessing stage. For evaluation purposes, we intend to develop new benchmarks specifically designed to assess the reasoning capabilities of MLLMs.